# THE ADDED VALUE OF MRI RADIOMICS AND DEEP LEARNING FOR GLIOBLASTOMA PROGNOSTICATION COMPARED TO CLINICAL AND MOLECULAR INFORMATION

Daniel Abler[1,2,3]*, Orso Pusterla[4]*, Anja Joye-Kühnis[4],
Nicolaus Andratschke[4], Michael Bach[6], Andrea Bink[7], Sebastian M. Christ[4], Patric Hagmann[8], Bertrand Pouymayou[4,7], Emanuele Pravatà[9,10,11], Piotr Radojewski[12], Mauricio Reyes[13,14], Lorenzo Ruinelli[20,21], Roger Schaer[3,15], Bram Stieltjes[15], Giorgio Treglia[10,18,19], Waldo Valenzuela[12], Roland Wiest[12], Solange Zoergiebel[2,17], Matthias Guckenberger[4], Stephanie Tanadini-Lang[4]**, Adrien Depeursinge[2,3,15]**

* shared first authorship, ** shared last authorship

[1]Department of Oncology, Geneva University Hospitals and University of Geneva, Geneva, Switzerland

[2]Lundin Family Brain Tumour Research Centre, Lausanne University Hospital, Lausanne, Switzerland.

[3]Institute of Informatics, University of Applied Sciences Western Switzerland (HES-SO), Sierre, Switzerland

[4]Department of Radiation Oncology, University Hospital Zurich, University of Zurich, Zurich, Switzerland

[5]Unit for Radiation Protection and Medical Physics, Cantonal Hospital Aarau, Aarau, Switzerland

[6]Department of Radiology and Nuclear Medicine, Basel University Hospital and University of Basel, Basel

[7]Department of Neuroradiology, Clinical Neuroscience Center, University Hospital Zurich, University of Zurich, Zurich, Switzerland

[8]Service of Diagnostic and Interventional Radiology, Lausanne University Hospital and University of Lausanne, Lausanne, Switzerland

[9]Division of Neuroradiology, Neurocenter of Southern Switzerland, Ente Ospedaliero Cantonale, Lugano, Switzerland.

[10]Faculty of Biomedical Sciences, Università della Svizzera Italiana, Lugano, Switzerland

[11]Department of Neuroscience, Imaging and Clinical Sciences and Institute for

Advanced Biomedical Technologies (ITAB), G. d'Annunzio University of Chieti-Pescara, Italy

[12]Institute for Diagnostic and Interventional Neuroradiology, Inselspital Bern, Bern, Switzerland

[13]ARTORG Center for Biomedical Engineering Research, University of Bern, Bern, Switzerland

[14]Department of Radiation Oncology, University Hospital Bern, University of Bern, Switzerland

[15]Department of Nuclear Medicine and Molecular Imaging, Lausanne University Hospital and University of Lausanne, Lausanne, Switzerland

[16]Department Research and Analytics, D&ICT, Basel University Hospital, Basel

[17]Department of Information and Communication Technology, Lausanne University Hospital and University of Lausanne, Lausanne, Switzerland

[18]Imaging Institute of Southern Switzerland, Ente Ospedaliero Cantonale, Bellinzona, Switzerland

[19]Faculty of Biology and Medicine, University of Lausanne, Switzerland

[20]Clinical Trial Unit, Ente Ospedaliero Cantonale, Bellinzona, Switzerland

[21]Team Innovation and Research, Area ICT, Ente Ospedaliero Cantonale, Bellinzona, Switzerland

ABSTRACT

**Background:**
Radiomics shows promise in characterizing glioblastoma, but its added value over clinical and molecular predictors has yet to be proven. This study assessed the added value of conventional radiomics (CR) and deep learning (DL) MRI radiomics for glioblastoma prognosis (≤6 vs > 6 months survival) on a large multi-center dataset.

**Methods:**
After patient selection, our curated dataset gathers 1152 glioblastoma (WHO 2016) patients from five Swiss centers and one public source. It included clinical (age, gender), molecular (MGMT, IDH), and baseline MRI data (T1, T1 contrast, FLAIR, T2) with tumor regions. CR and DL models were developed using standard methods and evaluated on internal and external cohorts. Sub-analyses assessed models with different feature sets (imaging-only, clinical/molecular-only, combined-features) and patient subsets (*S-1*: all patients, *S-2*: with molecular data, *S-3*: IDH wildtype).

**Results:** The best performance was observed in the full cohort (*S-1*). In external validation, the combined-feature CR model achieved an AUC of 0.75, slightly, but significantly outperforming clinical-only (0.74) and imaging-only (0.68) models. DL models showed similar trends, though without statistical significance. In *S-2* and *S-3*, combined models did not outperform clinical-only models. Exploratory analysis of CR models for overall survival prediction suggested greater relevance of imaging data: across all subsets, combined-feature models significantly outperformed clinical-only models, though with a modest advantage of 2–4 C-index points.

**Conclusions:** While confirming the predictive value of anatomical MRI sequences for glioblastoma prognosis, this multi-center study found standard CR and DL radiomics approaches offer minimal added value over demographic predictors such as age and gender.

## Key Points

Radiomics and clinical features predict glioblastoma survival, but in this large multi-center dataset, standard CR and DL radiomics from T1, T1 contrast, FLAIR, and T2 MRI added little prognostic value for 6-months survival beyond basic demographic information (age and gender).

## Importance of the Study

This study evaluated the added value of conventional (CR) and deep learning (DL) radiomics over clinical and molecular features for identifying high-risk glioblastoma patients with survival under 6 months. Using one of the largest multi-center glioblastoma imaging datasets to date and a rigorous validation framework, the analysis showed that adding radiomics to baseline demographic features (age, gender) led to only minimal improvements in predictive performance. These findings challenge the clinical utility of standard

CR and DL radiomics methods for glioblastoma risk stratification. While previous studies have reported added value of radiomics for risk or overall survival (OS) prediction, most relied on smaller, more homogeneous cohorts, and used less stringent evaluation approaches entailing risks of overfitting.

Moreover, they often lacked statistical comparisons of model performance. Differences in survival endpoints and evaluation metrics across studies suggest that imaging may still offer value for other clinical endpoints, warranting further investigation.

# 1 Introduction

Glioblastoma is the most frequent and most malignant primary brain cancer in adults, accounting for about 50% of malignant central nervous system tumors cases [1]. It infiltrates surrounding tissue, grows rapidly, and is frequently accompanied by compression and displacement of the surrounding tissue. Surgical resection to the maximal safe extent is the primary treatment for newly diagnosed glioblastoma, followed by radiation therapy alone or in combination with temozolomide (TMZ) chemotherapy and consolidation temozolomide [2]. The patient's age and Karnofsky performance status (KPS) are the most important clinical prognostic factors. Despite international research efforts, glioblastoma patients continue to have a poor prognosis, with a median survival of 12 to 14 m and a 5 y survival rate of 5.5 % [3], [4]. In addition to patient age and KPS, treatment selection relies on prognostic molecular markers, namely MGMT promoter hyper-methylation [2].

About 20% of patients already progress and eventually die during the 6-months standard-of-care treatment period [5]. Identification of these high-risk patients would be beneficial as it could help spare them the burden of a potentially quality-of-life consuming treatment.

Over the past decade, radiomics, the field of quantitative image-derived biomarker research, has attracted increasing attention. Based on the rationale that biomedical images contain information about the underlying pathophysiology, radiomics aims to capture the relevant characteristics via quantitative image analysis to predict clinically action-able outcomes (e.g., patient prognosis) [6]. Conventional radiomics (CR) and deep learning (DL) approaches have been investigated to address a wide range of questions in neuro-oncology [7], particularly linked to the clinical management of glioma and glioblastoma [8], [9], [10], including prognostic models for survival prediction and patient stratification [11]. An overview of the applications and potential of radiomics for precision medicine in this context is provided in [12].

While methodology-focused studies frequently investigate the predictive value of radiomics (i.e., image) information alone, many clinically oriented studies aim to integrate image-derived information with established clinical (e.g., age, performance status, extent of resection) or molecular (e.g., MGMT promoter hypermethylation) prognostic factors to maximize the prognostic value and evaluate the relative contribution of different types of information. However, despite numerous studies, the available evidence remains weak [13], mainly due to methodological issues, small sample size, and the lack of external validation cohorts [14].

The purpose of this study is to evaluate the prognostic value of pre-treatment (baseline) clinical and image-derived information for predicting 6-months survival in a large and diverse multicentric cohort of glioblastoma patients to identify patients with very short overall survival, which might benefit from quality-of-life centric treatment strategies including best supportive care. We compare the predictive performance of CR and DL models, employing standard modeling approaches, based on different combinations of image and clinical information. This allows us to investigate whether image-derived information provides added value compared to baseline clinical information in this prediction scenario.

## 2   Materials & Methods

### 2.1   Patient Population, Imaging & Clinical Datasets

This study builds on one of the largest imaging datasets for glioblastoma patients (WHO 2016), consisting of an initial set of 1030 patients treated across five hospitals in Switzerland between January 2004 and July 2021 and collected in the context of the SPHN-IMAGINE[1] project. The dataset contains depersonalized brain MR imaging of glioblastoma patients, as well as demographic, treatment, and molecular information. Data collection and analysis for this retrospective study were approved by the responsible ethics committee (BASEC-Nr. 2020-00859).

As part of their clinical diagnostic work-up, patients underwent MR imaging, including structural multi-contrast (or multi-parametric) MRI sequences: native T1-weighted (T1), post-contrast T1-weighted (T1c), T2-weighted (T2), and T2 fluid-attenuated inversion recovery (FLAIR) imaging. Imaging was conducted across multiple medical centers, utilizing scanners with 1.5T or 3T field strength and varying in manufacturer and model. The MR data were collected by the five treating hospitals, where patients underwent standard-of-care treatment, following personalized protocols according to each patient's best interest. Individual treatments involved a combination of biopsy or surgical resection of the tumor, radiotherapy, and concomitant or adjuvant chemotherapy, mainly utilizing either Temozolomide or Bevacizumab. Across all centers, a minimum set of patients' clinical, demographic, and treatment information was recorded, including the patient's age at diagnosis, sex, and overall survival (OS). Additionally, information about the tumor's MGMT promoter methylation status and IDH mutational status was collected for a subset of patients. However, as most data has been collected prior to 2021 (5th edition of WHO classification of CNS tumors), IDH mutational information was not available for all patients. We will therefore use the terminology of the WHO 2016 guidelines that defines glioblastoma without consideration of the tumor's IDH mutation status.

Inclusion in this study required the presence of an MR imaging examination including T1, T1c, T2, and FLAIR contrasts within 61 days before to 30 days after diagnosis and before any treatment (surgical resection, chemotherapy, or radiotherapy), as well as the availability of information related to patient age, sex, and the patient's survival at 6 months. Supplementary section 1.1 provides a detailed description of the MR sequence identification and selection approach. We found a set of 734 patients meeting inclusion criteria.

This dataset was further extended by an initial set of 611 patients from the public UPenn-GBM dataset [15] whose cohort characteristics and collected information closely align with the SPHN-IMAGINE project. 418 of the 611 UPenn-GBM patients met inclusion criteria. Hence, a total of 1152 patients were included in the study.

### 2.2   Image Data Preparation

All MR images were processed according to the workflow depicted in supplementary fig. 1: First, patients with complete pre-operative imaging datasets containing all four MRI sequences (T1, T1c, T2, FLAIR) were selected and DICOM images were converted to the NIfTI format. We used the automatic gliobloastoma segmentation software Deep-BraTumIA[2] (research version of [16]) to segment the tumors present in each patient's imaging dataset into regions of interest (ROIs) for the non-enhancing tumor (NET), enhancing tumor (ET) and edema, as well as for brain tissue. Exploiting the availability of these ROIs, the images' intensity levels were aligned by *z*-score standardization, correcting for the presence of tumor regions [17]. The processed images and segmentations were manually reviewed to ensure the inclusion of high-quality imaging exams with full brain coverage and all four MR contrasts. Sections 1.1 to 1.5 in the supplementary materials provide further details about image selection, segmentation, standardization, and review. The flow chart in fig. 1 summarizes the inclusion and exclusion criteria for this study.

---

[1] https://sphn.ch/network/projects/completed-projects_tiles/project-page_imagine/, as of May 2025.

### 2.3 Model Training & Evaluation Strategy

This study investigated both CR and DL approaches for distinguishing short (≤6 m) from long survivors (>6 m) based on baseline MR imaging, and in combination with clinical information such as age, sex, MGMT, and IDH status. These ML tasks address the clinically relevant scenarios detailed in Section 2.4. To evaluate the added value of imaging information for this prediction task, we compared the predictive performance of different models using either imaging or clinical information alone, or a combination of both.

Our evaluation strategy has two aims: First, to train and select optimally performing models, and to capture statistically significant differences in model performance while accounting for performance uncertainty linked to the choice of train/validation/test cohorts. Second, to verify the generalizability of these findings by model evaluation on an external test cohort. Figure 2 illustrates the training and evaluation strategy applicable to both CR and DL approaches and all outcomes of interest:

To implement this evaluation strategy, we divided our dataset into two subsets by centers, a cohort for model training and statistical evaluation (training set, which we further refer to as *TrainEval*, ~2/3 of the patients), and an external testing and replication cohort (*ExtTest*, ~1/3 of the patients). To ensure that patients in either cohort had been treated by different centers, we allocated patients from *SPHN-1*, *SPHN-2* and *UPENN* into the *TrainEval* cohort, and patients from *SPHN-3*, *SPHN-4*, *SPHN-5* into the *ExtTest* cohort. The *TrainEval* cohort was further divided into a *training* set (80 %) for hyperparameter selection and model training, and an *Internal Test set* (20 %) used for model evaluation. Model training and evaluation were repeated using $N$ different random data splits into *training* and *Internal Test set* to capture the influence of variability in both data subsets on model performance. Following statistical evaluation on the *TrainEval* cohort, a final model was evaluated on the *ExtTest* cohort. For CR analysis, we identified an optimal feature set and retrained a new model on the entire *TrainEval* cohort which was then evaluated on the *ExtTest* cohort. Evaluation of DL performance relied on ensembling the *ExtTest* predictions of the $N$ individual models trained in each *TrainEval* split. Further details about the modeling and evaluation approaches are described in supplementary section 2.1 and section 2.2 for CR and DL, respectively.

### 2.4 Clinical Prediction Scenarios, Cohort Subsets & Models
We distinguish three stages along a patient's clinical journey in this study based on the availability of clinical, imaging, and molecular information. Stage **A**: before the MR imaging exam (clinical information only), stage **B**: after imaging but before biopsy or surgery (clinical & imaging information), and stage **C**: after biopsy or surgery (clinical, imaging & molecular information). While stages **A** and **B** are non-invasive, stage **C** involves an invasive procedure (biopsy or surgical resection) that provides samples of tumor tissue, thus enabling the tumor's molecular characterization. However, some patients may not reach **C** as their overall health condition prevents invasive procedures, or they may choose not to undergo such procedures. Consequently, for this subgroup of patients, the information obtained by analyzing samples of tumor tissues will not be available. This might include information about important diagnostic-, prognostic-/predictive- markers, such as *IDH1/2* mutation status or *MGMT* hyper-methylation status, respectively.

---
[2] https://www.nitrc.org/projects/deepbratumia/, as of May 2025.

**Cohort Subsets & Models** For investigating 6-months survival prediction, we distinguish three cohort subsets depending on the availability of molecular information and the patients' IDH status:

- **Subset *S-1*** includes all patients with available MR data, and age and sex information.
- **Subset *S-2*** includes all patients of subset *S-1* with available MGMT status (*methylated* vs. *unmethylated*).
- **Subset *S-3*** includes all patients of subset *S-2* with *wildtype* IDH status, i.e., focusing on primary glioblastoma as defined in the WHO 2021 classification.

For each subset, we investigated models using demographics (*age*, *sex*, i.e., stage **A**), image-derived (*radiomics*, i.e., stage **B**: both CR or DL) and molecular features (*mgmt*, i.e., stage **C**) alone and in combination to assess the contribution of each feature group to prediction performance. Table 1 summarizes all different models and their respective data and feature sets.

### 2.5 Conventional Radiomics and Deep Learning Modeling

While relying on the same general model training and evaluation strategy described in section 2.3, the modeling approaches for CR and DL radiomics were designed to leverage distinct workflows tailored to their respective methodologies. For CR, this involved the extraction and selection of 107 standardized hand-crafted radiomics descriptors from each MR modality (T1, T1c, T2, FLAIR) and for each of the six primary and combined ROIs (NET, ET, edema, ET-edema, NET-ET, NET-ET-edema), as well as training and validation of conventional ML models. For DL, all standardized images were aligned by rigid registration to the MNI152 (T1, 1mm×1mm×1mm) atlas[3] and cropped to contain the minimum useful field of view obtained by superposing all brain masks in the atlas frame. We employed a neural network architecture optimized for both image and clinical parameter integration. Models including either imaging or clinical features were evaluated separately.

Supplementary material provides comprehensive details on the CR and DL modeling methodology. See supplementary sections 1.6 and 1.7 for in-depth descriptions of the CR-feature extraction process, as well as for CR and DL-based modeling.

### 2.6 Statistical Methods

Differences in variable distributions between *TrainEval* and *ExtTest* cohorts were assessed via Student's t-test or Mann-Whitney U-test as appropriate for continuous variables, via proportions *z*-test for binary variables, and via log-rank test for time-to-event data. We interpreted p-values $p \leq 0.05$ to indicate statistical significance.

---

[3] https://nist.mni.mcgill.ca/icbm-152lin/, as of May 2025.

Computational data analysis was performed on the Leonhard Med[4] secure trusted research environment at ETH Zurich. The SPHN IMAGINE server configuration provided access to 34 Intel Xeon Cascadelake CPUs running at 2.6GHz, and eight NVIDIA RTX 2080 GPUs with 11GB of RAM.

## 3 Results

### 3.1 Dataset for Modeling

The initial dataset included a total of 1641 glioblastoma patients, with varying availability of demographic and clinical information, as well as MRI imaging. Of these, 1030 glioblastoma patients were collected as part of the SPHN study, while 611 were retrieved from the UPENN study. For 1208 patients (780 from SPHN and 428 from UPENN) complete multi-contrast MRI imaging as well as information regarding 6-months survival and demographic details were available. After manual review of the MR imaging data, the final cohort included in this study and used for 6-months survival prediction comprised a curated dataset of 1152 glioblastoma patients, 734 from SPHN and 418 from UPENN. Table 2 summarizes the number and demographic characteristics of the patients included in this study, divided by subset. No statistically significant differences were found between the distributions of clinical covariates (age, sex, mgmt status) and outcome (6-months survival) in the *TrainEval* and *ExtTest* cohort for all data subsets except for patient age in subset *S-1* ($p = 0.001$).

Further information is provided in the supplementary material: Table 1 details demographic information for the individual datasets from each center (*SPHN-1* to *SPHN-5*, *UPENN*). Table 2 summarizes the statistical tests performed and their results. Figure 4a compares the survival patterns of the various cohorts across data subsets, figures 4b to 4d illustrate survival patterns for each data subset by data origin.

### 3.2 Predictive Performance of Models

Table 3 presents the Area under the Curve (AUC) of the Receiver-Operating Characteristics (ROC) for the conventional radiomics and DL models and across the three investigated subsets. Supplementary fig. 5 presents a visual comparison of statistical evaluation performance (Internal Test) on the *TrainEval* cohort.

**Conventional Radiomics Performance** Prediction performance on the *TrainEval* cohort reached a maximum mean AUC of 0.81 and 0.75 in cross-validation (CV) and on the *Internal Test* set, respectively. Similarly, performance on the unseen *ExtTest* cohort also reached up to 0.75 AUC. In subsets *S-1* and *S-2*, models combining clinical and radiomics information achieved the best *ExtTest* performance. However, differences in mean *Internal Test* performance across the various models within the same data subset were small and, in most cases, not statistically significant, supplementary fig. 6a. An exception is subset *S-1* where the model containing both demographics and clinical information, *M1-demo-imag*, performed significantly better ($p = 0.002$) than model *M1-demo*, relying exclusively on demographic information.

The performance of equivalent models differed across data subsets, typically deteriorating with dataset size. For example, *S-1* models *M1-img* and *M1-demo-img* achieved significantly higher AUC than their counterparts on subset *S-2* (*M2-img*, *M2-demo-img*) and subset *S-3* (*M3-img*, *M3-demo-img*). Similarly, although not statistically significant, mean *Internal Test* performance of *S-2* models *M2-demo-mgmt* and *M2-demo-mgmt-img* exceeded the mean *Internal Test* performance of their counterparts on subset *S-3* (*M3-*

---
[4] https://sis.id.ethz.ch/services/sensitiveresearchdata , as of May 2025.

*demo-mgmt* and *M3-demo-mgmt-img*). Addition of MGMT information to models in subsets *S-2* and *S-3* did not affect prediction performance in a statistically significant way, compare e.g., models *M2-demo* to *M2-demo-mgmt* or *M2-demo-img* to *M2-demo-mgmt-img*.

Performance on the unseen *ExtTest* cohort was compatible with (within 95% CI) or exceeded performance estimates on the *Internal Test* set across models and data subsets, with model *M1-img* being the only exception.

**Deep Learning Performance** Performance reached a maximum AUC of 0.71 on the cross-validation set, 0.70 on the *Internal Test* set, and 0.74 on the *ExtTest*. As expected, CV performances surpassed those of the Internal Test set, while performance on *ExtTest*, leveraging ensemble averaging, outperformed both CV and Internal Test performances. In contrast to the CR evaluation, no statistically significant difference was observed between model *M1-demo-imag* and model *M1- demo*, supplementary fig. 6b.

Overall, performance declined from data subset *S-1* to *S-2* and *S-3*, which we attribute to the reduced cohort sizes hindering the training of the DL model. Demographic-only models achieved the best performances, outperforming image-only models (e.g. 0.74 for *M1-demo*, and 0.70 for *M1-img*, on the *ExtTest* cohort). Surprisingly, image or demographic-image models tended to show a slight but significant decline in performance as compared to demographic-only models, for example between *M2-img* and *M2-demo* ($p = 0.017$), between *M3-demo-img* and *M3-demo* ($p=0.014$). Similarly, models leveraging MGMT information in subsets *S-2* and *S-3* tended to exhibit a slight performance decrease. Furthermore, the results for subset *S-2* appeared superior to those of *S-3* (patients with confirmed IDH wildtype), which may partially be attributed to the differences in patient numbers.

**Conventional Radiomics vs. Deep Learning** We found an overall decrease of internal and external test performances from data subset *S-1* to *S-2* and from *S-2* to *S-3* for both CR and DL radiomics. As expected, the performance of models utilizing only clinical information (*M{1,2,3}-demo*, *M{1,2,3}-demo-mgmt*) agreed very well across CR and DL, despite the different modeling approaches. CR models using only image or both image and clinical features out-performed their DL counterparts in the statistical evaluation on subsets *S-1* and *S-2* ($p < 0.05$). In subset *S-3* these differences were not statistically significant. While this advantage was not reflected in the *ExtTest* cohort for image-only models, it held for models using both image and clinical features. Overall, the performance of these *fusion* models combining image and clinical information, was more closely aligned to the performance of clinical-only models in the case of CR, but to image-only models in the case of DL.

## 4 Discussion

### 4.1 Summary

This study investigated the feasibility of predicting the 6-month survival in newly diagnosed glioblastoma patients using base-line demographic (*age*, *sex*), molecular (*MGMT* hyper-methylation status), and image-derived (MR-scans at baseline) information. We comprehensively investigated both conventional radiomics (CR) and deep learning (DL) models on one of the world's largest glioblastoma imaging datasets, comprising curated data of 1152 patients.

Across all data subsets and modeling approaches, demographics-only models (*age*, *sex*) performed best or on-par with more complex models, reaching an AUC of 0.74 on the *ExtTest* cohort. Older age by itself is a well-known and major prognostic factor for glioblastoma patients [18]. Models using imaging information alone generally performed worse (up to AUC 0.70 for DL, 0.68 for CR in subset *S-1*). A small but statistically significant (p = 0.022) added value of imaging information over demographic baseline information for prediction on the *ExtTest* cohort prediction was observed in subset *S-1* using CR (AUC 0.75). MGMT status, although associated with improved overall survival (OS) in our cohort, was not significantly associated with 6-month survival (chi-squared test p = 0.4).

While we found imaging information to provide a slight added value over baseline clinical models using CR, this added value could not be reproduced by DL, likely due to its early-fusion approach (direct feature concatenation). In contrast, the CR approach employed late-fusion, combining imaging-derived radiomics risk scores (RRS) with demographic data in a dedicated linear model. In DL models, the combined predictions (AUC 0.74) were more correlated with imaging-only predictions (Pearson correlation coefficient r = 0.89) than with demographic-only predictions (r = 0.60), suggesting underuse of demographic information. Attempts to mimic late fusion in DL by limiting the number of imaging features did not improve results. As noted in [18], [19], imaging features might add independent prognostic factors, distinct from traditional prognostic variables. Consequently, the integration of imaging and demographic data may have introduced complementary but conflicting information and noise, potentially hindering the model's ability to learn effectively from the combined data.

Despite achieving a peak AUC of 0.75 (CR) and 0.74 (DL) on *ExtTest* for 6-month prediction, such performance levels (AUC <0.8) may not suffice for clinical decision-making. Nevertheless, the models allowed stratification into subgroups with significantly different survival outcomes (see Supplementary Fig. 10).

## 4.2 Extension to Overall Survival

To assess the generalizability of the CR modeling results to alternative prediction tasks, we evaluated model performance on overall survival (OS) prediction using a survival analysis framework—specifically, the Cox proportional hazards model—rather than a classification approach. Using the same features and datasets, we observed that models incorporating imaging data consistently outperformed clinical-only models across all data subsets, as measured by concordance index (C-index). Although the improvements were statistically significant, the numeric gains were modest, with imaging-enhanced models showing an advantage of approximately 2–4 C-index points — comparable to the gains observed in the 6-month survival prediction task. For instance, in subset S1, models M1-demo, M1-img, and M1-demo-img achieved C-index values of 0.64 [0.63, 0.66], 0.66 [0.64, 0.67], and 0.68 [0.67, 0.69], respectively.

In contrast to the 6-month prediction scenario, the inclusion of MGMT status further improved OS prediction, particularly when combined with imaging features. For example, in subset S3, the C-index increased from 0.62 [0.60, 0.63] for M3-demo to 0.65 [0.63, 0.66] for M3-demo-MGMT, and to 0.68 [0.67, 0.70] for M3-demo-MGMT-img. Full OS modeling results are presented in Supplementary Section

## 4.3 Comparison with prior studies

Two modeling strategies dominate the glioblastoma prognostic radiomics literature. Table 4 summarizes representative studies and their findings:

*Type A* follows a standard CR modeling workflow where model development and testing rely on separate datasets. Prognostic radiomics features are selected using a subset of the study cohort (corresponding to *TrainEval*, see Section 2.3), and a predictive model is then trained on this dataset using the selected radiomics features along with other clinical and molecular variables. The resulting model is evaluated on another dataset (test set) by predicting the patient's outcomes and measuring performance. By repeating this process with different combinations of variable sets, the added prognostic value of each information source can be evaluated. Employing this strategy, [20] reported that the addition of radiomics features to a model already using clinical and molecular characteristics improved OS prediction as measured by an increase in the integrated area under the time-dependent ROC curve (iAUC) from 0.70 [0.58,0.80] to 0.78 [0.69,0.87]. Similarly, [21] found that a combined model using clinical, molecular and radiomics information was better able to predict low-risk (LR) patients with OS≥18 m (AUC 0.89), compared to models based on *age* alone (0.70). Using advanced deep learning strategies, [22] demonstrated a marginal increase of C-index from 0.675 [0.639,0.711] to 0.688 [0.653,0.723] (*p*=0.052) when combining imaging and non-imaging features in a sub-cohort of glioblastoma patients. Similarly, [19] employed a state-of-the-art transformer-based deep learning model for nonlinear and non-proportional survival prediction in glioblastoma, showing almost no improvement in the time-dependent concordance index with the inclusion of imaging features (mean ± 95% CI: $0.669 \pm 0.021$ vs. $0.672 \pm 0.023$).

*Type B* combines predictive modeling techniques with methods commonly used for assessing clinical biomarkers. A scalar RRS, linked to OS, is developed on a subset of the study cohort (training) by applying prediction modeling techniques to imaging information. This model is then used to predict the RRS for patients of another cohort (test). The prognostic value of RRS alone and in combination with clinical, molecular, and genetic features or derived scores is assessed on the test cohort by multivariate regression analysis. Fitting a new model to the test set introduces a risk of overfitting that may result in over-estimation of the overall prognostic value, particularly when non-linear models are used. Using hand-crafted and DL radiomics features from a cohort of 404 patients, [23] developed an RRS to distinguish high-risk patients with OS ≤ 6 m from low-risk patients. Multivariate survival analysis using different feature sets and the predicted RRS from a replication cohort of 112 patients, showed an improvement in average performance metrics with every layer of information added to the clinical baseline model, reaching a C-index of up to 0.75 [0.72,0.79] when jointly considering clinical, RRS and molecular information. Instead of deriving the RRS from a classification model, [24], [25], [26], [27] developed survival models for RRS prediction. Similarly to [23], these studies employed multivariate survival analysis to estimate hazard ratios and the prognostic performance of different combinations of clinical and molecular variables, as well as the RRS.

In summary, *Type A* studies employ a stricter validation strategy than *Type B*, the latter relying on both refitting, as well as evaluating, the model on the entire dataset. Nevertheless, the risk of overfitting in *Type B* studies remains moderate since the final model typically uses fewer than 10 features.

The present study employed a very strict *Type A* model validation strategy. Models corresponding to different clinical scenarios (section 2.4) and utilizing different sets of information were trained and evaluated on two distinct cohorts: *TrainEval* including an *Internal Test* set for statistical evaluation and allowing the computation of average (internal) test performance estimates and their 95% CIs, as well as *ExtTest*, an external test set, consisting of patients from treatment centers different from those used for training and internal testing.

All studies in Table 4 found that adding image-derived information (individual features or RRS) improved prognostic performance, suggesting its added value over clinical and molecular data alone. However, only [20], [21] demonstrated a statistically significant benefit, and most other studies lacked formal testing, with overlapping confidence intervals suggesting no significant improvement. Some prior studies focused on different endpoints or had access to additional clinical variables not available in our dataset, such as EOR or KPS. For instance, [20] reported a top model achieving an AUC of 0.89 for identifying low-risk patients (OS > 18m), whereas this study targeted high-risk patients (OS ≤ 6m). Most others predicted OS with C-indices from 0.71 to 0.88.

While OS prediction was not the focus of this study, our CR clinical-image models (early-fusion, Type A evaluation) reached C-indices of up to 0.68 [0.67,0.70] and 0.67 on the internal and external test sets, respectively, supplementary table 5 and fig. 11. This performance aligns with recent studies using strict Type A evaluation on large, diverse datasets [19], [22], while the higher C-indices reported in other studies [24], [27] likely reflect the use of less-strict Type B evaluation approaches and access to additional clinical variables such as extent of resection (EOR), Karnofsky performance status (KPS), and treatment information not available in the SPHN dataset.

### 4.4 Limitations

While our study followed best practices for training and evaluating CR and DL models, several limitations exist: (1) The heterogeneous glioblastoma patient population, with varied treatments and outcomes, may impact model generalizability. (2) Although imaging data were manually curated and preprocessed, variability due to scanner type, vendor, and acquisition time (2004–2021) remains as is typical for real-world imaging data. While unsupervised clustering showed differences in feature distributions, especially between SPHN and UPENN datasets, batch harmonization (ComBat) had no significant effect on CR model performance, compare supplementary Table 4 (CR: with/without SMOTE, with batch harmonization) to Table 3 (CR: with SMOTE, without batch harmonization). (3) From a methodological perspective, we focused on conventional modeling techniques using image-derived and basic clinical features. For DL, we tested various architectures including 2.5D and 3D models with different input configurations (e.g., stacked MR data, tumor segmentation, multi-channel approaches), varied network parameters (e.g., depth, kernel sizes, channels), optimized hyperparameters, and applied data augmentation. While these approaches were well-optimized, we acknowledge that more advanced techniques—such as attention-based architectures or pre-trained DL models using autoencoders—may further improve performance. Similarly, CR models could benefit from incorporating domain-specific features like the VASARI set [25] or advanced post-processing techniques that capture tumor effects on surrounding brain structures, including brain parcellation [28] and deformation modeling [29], [30]. Additionally, including clinical covariates such as performance status and extent of resection—shown to be prognostically valuable in prior studies [20], [23], [24], [26], [27], [31]—could enhance prediction but were not available in the SPHN-imagine dataset.

### 4.5 Conclusion

This study, based on a large multi-center dataset of 1,152 glioblastoma patients and a rigorous validation approach, showed that image-based models can distinguish between high- and medium/low-risk groups with significantly different survival outcomes. However, the added predictive value of imaging over basic demographic factors (age, gender) for predicting 6-month survival was small and not consistently reproducible across different models and patient subgroups. These findings challenge the utility of routine imaging information in combination with standard CR and DL radiomics modeling approaches for glioblastoma risk stratification. They also emphasize the need for large-scale real-world (multi-institutional) validation studies.

### Ethics

Collection and analysis of retrospective data in the SPHN-IMAGINE project were approved by the responsible ethics committee (BASEC-Nr. 2020-00859).

## Funding

This work was partially funded by the Swiss Personalized Health Network (SPHN) via the IMAGINE and QA4IQI projects, the Swiss Cancer Research foundation with the project TARGET (KFS-5549-02-2022-R); the Lundin Family Brain Tumour Research Centre at CHUV; the Hasler Foundation with the project MSxplain number 21042; and the Swiss National Science Foundation with projects 205320_219430 and 325230_197477.

## Conflict of Interests

The authors declare that they have no known competing financial interests or personal relationships that could have appeared to influence the work reported in this paper.

## Authorship

**Daniel Abler**: Conceptualization, Methodology, Software, Validation, Formal analysis, Investigation, Data Curation, Writing – Original Draft, Visualization; **Orso Pusterla**: Conceptualization, Methodology, Software, Validation, Formal analysis, Investigation, Data Curation, Writing – Original Draft, Visualization; **Stephanie Tanadini-Lang**: Conceptualization, Methodology, Validation, Formal analysis, Resources, Visualization, Supervision, Project administration; **Adrien Depeursinge**: Conceptualization, Methodology, Validation, Formal analysis, Resources, Writing – Original Draft, Supervision; **Roger Schaer**: Software; **Anja Joye-Kühnis**: Validation, Formal analysis, Investigation, Data Curation; **Emanuele Pravatà**: Data Curation; **Matthias Guckenberger**: Resources, Funding acquisition; **Waldo Valenzuela:** Data Curation; **All authors**: Writing – Review & Editing

## Data Availability

Code for data preparation and modeling, if not publicly available, will be shared with interested researchers or clinicians upon request. The SPHN datasets *SPHN-1* to *SPHN-5* are not currently publicly available, however, we are working towards the release of this resource. The *UPENN* dataset [15] is available from the Cancer Imaging Archive[5].

---

[5] https://www.cancerimagingarchive.net/collection/upenn-gbm/

**Figures & Tables (Main Manuscript – in order of occurrence in text)**

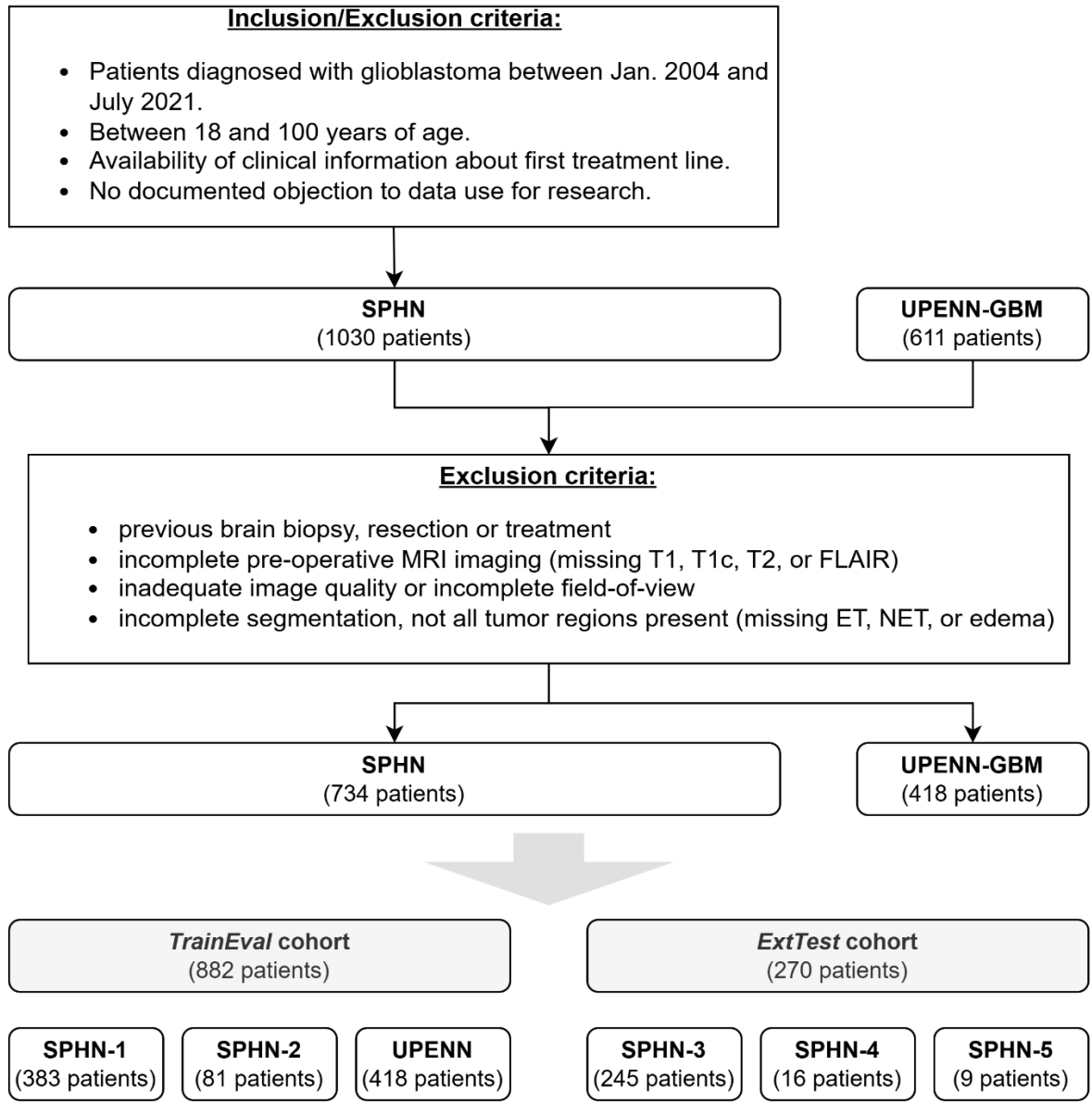

Figure 1: Inclusion and exclusion criteria for this study.

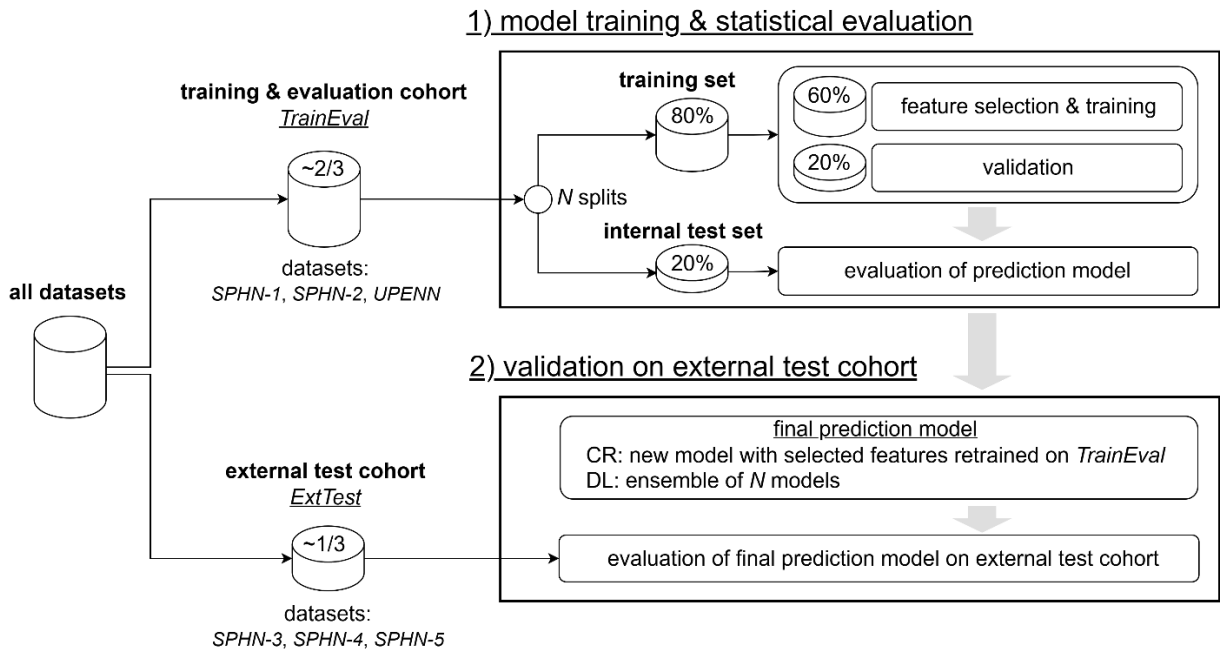

Figure 2: Model training-evaluation and testing strategy for CR and DL approaches: The dataset was divided into two cohorts for (1) model training and statistical evaluation (*TrainEval*: *SPHN-1*, *SPHN-2*, *UPENN*) and (2) external validation (*ExtTest*: *SPHN-3*, *SPHN-4*, *SPHN-5*). Within the *TrainEval* cohort, the dataset was randomly divided into training/validation and Internal Testing sets. This process was repeated *N* times to estimate prediction uncertainty. The final prediction model obtained from the *TrainEval* cohort was validated on the unseen *ExtTest* cohort.

| Model | Data Subset | Feature Sets |
|---|---|---|
| *M1-demo* | *S-1* | demographics |
| *M1-img* | *S-1* | image |
| *M1-demo-img* | *S-1* | demographics, image |
| *M2-demo* | *S-2* | demographics |
| *M2-img* | *S-2* | image |
| *M2-demo-img* | *S-2* | demographics, image |
| *M2-demo-mgmt* | *S-2* | demographics, molecular |
| *M2-demo-mgmt-img* | *S-2* | demographics, molecular, image |
| *M3-demo* | *S-3* | demographics |
| *M3-img* | *S-3* | image |
| *M3-demo-img* | *S-3* | demographics, image |
| *M3-demo-mgmt* | *S-3* | demographics, molecular |
| *M3-demo-mgmt-img* | *S-3* | demographics, molecular, image |

Table 1: Models investigated for different combinations of data subsets and features.

| Subset | Cohort | Patients | Age | | | Sex | MGMT | | Survival | | |
|---|---|---|---|---|---|---|---|---|---|---|---|
| | | | mean | median | range | female | meth | NA | mean | median | ≤6m |
| | | # | [y] | [y] | [y] | [%] | [%] | [%] | [d] | [d] | [%] |
| All | | 1152 | 64.1 | 65.0 | [19, 90] | 36.2 | 41.2 | 32.3 | 465.3 | 380.0 | 24.5 |
| S-1 | TrainEval | 882 | 63.5 | 64.0 | [19, 90] | 36.6 | 39.8 | 40.7 | 467.1 | 377.0 | 24.3 |
| | ExtTest | 270 | 66.3 | 67.0 | [25, 90] | 34.8 | 44.0 | 4.8 | 459.2 | 386.5 | 25.2 |
| S-2 | TrainEval | 523 | 65.5 | 66.0 | [21, 90] | 34.6 | 39.8 | 0.0 | 449.4 | 373.0 | 22.9 |
| | ExtTest | 257 | 66.1 | 67.0 | [25, 90] | 34.6 | 44.0 | 0.0 | 447.1 | 387.0 | 24.1 |
| S-3 | TrainEval | 456 | 65.4 | 66.5 | [21, 90] | 33.8 | 38.8 | 0.0 | 427.3 | 365.0 | 23.5 |
| | ExtTest | 244 | 66.5 | 67.0 | [25, 90] | 34.4 | 44.7 | 0.0 | 441.0 | 385.5 | 24.6 |

Table 2: Demographic characteristics of the glioblastoma patients resulting from manual revision and included in this study for modeling, divided by subsets and *TrainEval* vs. *ExtTest* cohorts. As the availability of *MGMT* status information is not required for subset *S-1*, we report both the percentage of patients with MGMT hyper-methylation (*meth*) and without MGMT information (*NA*). The percentage of patients with MGMT hyper-methylation refers to the group of patients with known MGMT status.

|  |  | **Conventional Radiomics** | | | **Deep Learning** | | |
|---|---|---|---|---|---|---|---|
|  |  | *TrainEval* | | *ExtTest* | *TrainEval* | | *ExtTest* |
|  |  | *CV* | *Internal Test* |  | *CV* | *Internal Test* |  |
| *S-1* | *M1-demo* | 0.69 [0.67,0.71] | 0.69 [0.67,0.72] | 0.74 | 0.70 [0.67,0.71] | 0.70 [0.67,0.72] | **0.74** |
|  | *M1-img* | 0.77 [0.76,0.78] | 0.72 [0.70,0.74] | 0.68 | 0.70 [0.68,0.72] | 0.67 [0.65,0.69] | 0.70 |
|  | *M1-demo-img* | 0.80 [0.79,0.81] | 0.75 [0.73,0.78] | **0.75** | 0.71 [0.70,0.73] | 0.67 [0.65,0.69] | **0.74** |
| *S-2* | *M2-demo* | 0.68 [0.67,0.69] | 0.66 [0.63,0.68] | **0.73** | 0.70 [0.68,0.73] | 0.65 [0.62,0.68] | **0.73** |
|  | *M2-img* | 0.77 [0.75,0.78] | 0.66 [0.63,0.69] | 0.65 | 0.68 [0.66,0.70] | 0.60 [0.56,0.63] | 0.69 |
|  | *M2-demo-img* | 0.79 [0.78,0.81] | 0.68 [0.64,0.71] | **0.73** | 0.68 [0.65,0.71] | 0.61 [0.58,0.65] | 0.70 |
|  | *M2-demo-mgmt* | 0.69 [0.67,0.70] | 0.68 [0.65,0.71] | 0.72 | 0.67 [0.65,0.70] | 0.65 [0.62,0.69] | **0.73** |
|  | *M2-demo-mgmt-img* | 0.81 [0.79,0.82] | 0.70 [0.66,0.73] | 0.72 | 0.66 [0.64,0.67] | 0.61 [0.58,0.64] | 0.69 |
| *S-3* | *M3-demo* | 0.67 [0.65,0.69] | 0.65 [0.62,0.68] | **0.73** | 0.66 [0.63,0.69] | 0.65 [0.62,0.68] | **0.72** |
|  | *M3-img* | 0.77 [0.75,0.78] | 0.62 [0.59,0.65] | 0.65 | 0.67 [0.64,0.69] | 0.60 [0.56,0.64] | 0.67 |
|  | *M3-demo-img* | 0.81 [0.79,0.83] | 0.64 [0.60,0.67] | 0.70 | 0.66 [0.63,0.68] | 0.58 [0.55,0.62] | 0.64 |
|  | *M3-demo-mgmt* | 0.67 [0.65,0.69] | 0.67 [0.64,0.71] | 0.71 | 0.63 [0.61,0.65] | 0.63 [0.60,0.67] | 0.69 |
|  | *M3-demo-mgmt-img* | 0.81 [0.79,0.83] | 0.66 [0.62,0.69] | 0.72 | 0.68 [0.65,0.71] | 0.61 [0.59,0.64] | 0.65 |

Table 3: Comparison of Area Under the ROC Curve (ROC-AUC) performances for all different conventional radiomics and DL models across the three data subsets *S-1*, *S-2*, *S-3*. Estimates for *CV* and *Internal Test* performance were obtained by statistical evaluation on the *TrainEval* cohort, *test* performance was estimated by evaluation on the unseen *ExtTest* cohort. Results from statistical evaluation are reported as mean [95% CI], results from evaluation on the *ExtTest* cohort correspond to single performance estimates.

| study | type | cohort size | | | outcome | | best performing model | | radiomics | |
|---|---|---|---|---|---|---|---|---|---|---|
| | | total | train | test | RRS | fusion | variable sets | performance | add. value | stat. sig. |
| this | A | 1152 | 882 [(m)] | 270 [(e)] | HR HR[DL] | HR HR[DL] | Clin *(age, sex)* + RRS Clin *(age, sex)* +/- RRS | AUC  0.75 AUC  0.74 | Y N | Y [†] N [†] |
| [20] | A | 217 | 163 | 54 | OS [‡] | | Clin *(age, EOR, KPS, treatment)*, Mol *(MGMT, IDH)*, Img *(18 features)* | iAUC  0.78 | Y | N [††] |
| [21] | A | 156 | 116 | 40 [(e)] | LR [‡] | | Clin *(age)*, Mol *(MGMT)*, Img *(7 features)* | AUC  0.89 | Y | Y [†††] |
| [22] | A[DL] | 1528[(g)] | 1085[(m)] | 443[(b,e)] | HR | OS | Clin *(age, sex, EOR, KPS)*, Mol *(MGMT, IDH)*, Img *(DL-index)* | C-index  0.69 | Y | N [††††] |
| [19] | A[DL] | 744 | 378[(m)] | 366[(e)] | OS | OS | Clin *(age, sex, EOR, KPS)*, Mol *(MGMT, IDH)*, Img *(DL-features)* | time-dependent C-index  0.67 | Y | N [††] |
| [23] | B | 516 | 404 | 112 | HR vs. LR[DL] | OS | Clin *(age, sex, EOR)*, Mol *(MGMT)* + RRS | C-index  0.75 | Y | N [††] |
| [27] | B | 313 | 131 [(m)] | 182 [(e)] | OS | OS | Clin *(age, EOR)*, Mol *(MGMT, IDH)* + RRS | C-index  0.88[§] | Y | n.a. |
| [25] | B | 188 | 142 | 46 | OS | OS | Clin *(age, sex, surgery, treatment)*, Mol *(MGMT)* + VASARI score, RRS | C-index  0.71 | Y | N [††] |
| [24] | B | 187 | 122 [(m)] | 65 [(e)] | OS | OS | Clin *(age, KPS, treatment)* + RRS [‖] | C-index  0.78, 0.79[‖] | Y | n.a. |
| [26] | B | 181 | 120 | 61 | OS | OS | Clin *(age, KPS, EOR, treatment)*, Mol *(MGMT)* + RRS | BrSR  -37% | Y | Y [††††] |

(m) Multi-institutional dataset, i.e., data acquired by multiple institutions.
(e) External test set, potentially multi-institutional.
(g) Training on a glioma (low and high grade) cohort, testing on Glioblastoma.
(b) Two glioblastoma sub-cohorts.
DL: Using DL approach instead of CR.
[†] Statistical significance based on permutation test on bootstrapped test results.
[††] Statistical significance based on 95% CI of Cox-Model regression fit (i.e. no overlap of CIs).
[†††] Statistical significance based on ANOVA.
[††††] Statistical significance based on Student's t-test for dependent samples.
[†††††] Statistical significance based on DeLong test.
[‡] Only radiomics feature selection was performed on training data, no RRS prediction model was trained. To evaluate the prediction performance of different sets of variables, a new model was trained using the selected radiomics features along with other variable sets to be tested.
[§] Added value of radiomics observed only in sub-population of male patients.
[‖] Various RRS were investigated.

Table 4: Unless indicated otherwise, the studies used CR approaches for building the radiomics risk score (RRS) and/or evaluating the combined performance of multiple feature groups. High risk (HR) corresponds to overall Survival (OS) of ≤6 m, low risk (LR) to OS >18 m. Clinical (Clin), Molecular (Mol) and Image-derived (Img) feature groups may consist of one or more variable, such as *age*, *sex*, extent of resection (*EOR*), Karnofksy performance status (*KPS*), treatment information, MGMT methylation status (*MGMT*), IDH expression (*IDH*), VASARI imaging features, individual radiomics features, or a model-derived radiomics risk score (RRS). Model performances were evaluated by the area under the ROC curve (AUC) for classification outcomes, and the (time-dependent) concordance index (C-index), integrated AUC (iAUC) or reduction in Brier Score relative to a baseline model (BrSR) for survival outcomes. Statistical significance (stat. sig.) was assessed at the $p = 0.05$ threshold. The table provides a concise overview of key findings relevant to our research from various complex studies; for a more comprehensive understanding, the authors encourage reading the full papers of the respective studies.

# Supplementary materials:

# "The added value of MRI radiomics and deep learning for Glioblastoma prognosis compared to clinical and molecular information"

## 1 Imaging Data Preparation

Images were processed according to the workflow depicted in fig. 1. Section 1.1 details all individual steps involved in image selection, review, and processing.

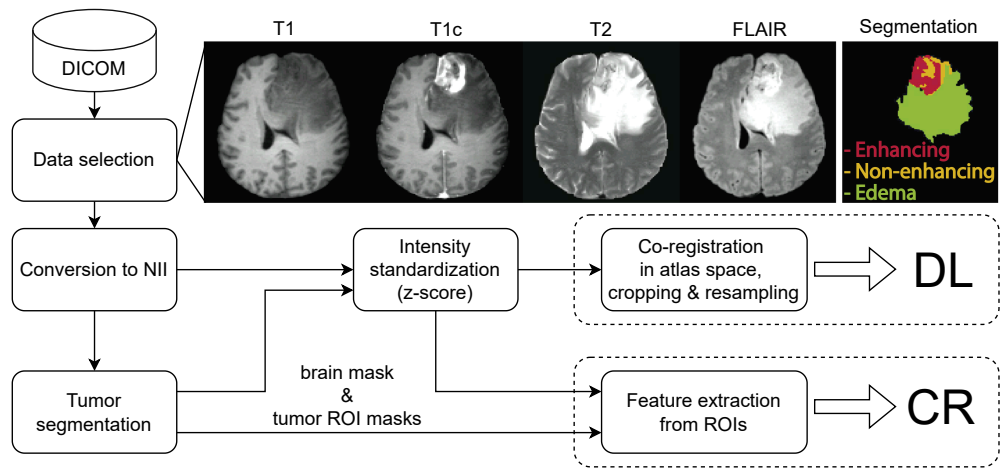

Figure 1: Image processing workflow. Major steps include the selection of suitable DICOM files, conversion, tumor segmentation, and intensity standardization. Radiomics features extracted for all ROIs and from all modalities were used for CR modeling. DL modeling relied on intensity-standardized images, co-registered into a shared frame-of-reference.

### 1.1 Automated MR Image Selection

To identify patients with complete pre-operative imaging datasets, the most relevant MR imaging studies were selected and available MR sequences were labeled as T1-weighted (*T1*), T1-weighted post-contrast (*T1c*), T2-weighted (*T2*) and *FLAIR* images using a semi-automatic rule-based approach relying on DICOM metadata and tailored to the respective dataset. First, all MR imaging studies in the period between 61 days before and up to 30 days after the patient's diagnosis were considered. Then, non-brain imaging studies and imaging that occurred after the first treatment (surgery, systemic treatment or radiation therapy) were discarded. From the remaining studies, the study closest to the diagnosis date was selected. When multiple instances of a specific MR contrast were available at the selected imaging timepoint, one single high-quality MR series was selected by prioritizing higher resolution (smaller maximum spacing in any direction) and axial acquisitions over other acquisition modes. Any remaining selection ambiguities were resolved by selecting the *most representative* MR series for the respective MR modality and dataset. This selection relied on quality metrics computed by the MRQY[1] tool. Specifically, *contrast per pixel* (CPP), *coefficient of variation percent* (CV), *mean* (MEAN) and *range* (RNG) of foreground intensity values, as well as *signal-to-noise ratios* (SNR1, SNR2) were computed for all selected series of the respective MR contrast, as well as for the series that remained to be selected. For each candidate series for a specific patient and MR contrast, the Root Mean Square (RMS) was computed from these metrics, and the candidate series with the smallest RMS value was selected. Patients for whom only a subset of the required MR contrasts was available were not included in the final dataset.

## 1.2 Manual MR Image Review

The MRI data of this dataset were manually reviewed and about 4.6% of the MR data had to be removed due to one or more of the following reasons: 2.9% exhibited suboptimal diagnostic image quality attributed to motion, artifacts, or insufficient resolution; 0.7% had segmentation flaws, or the segmentation did not include the ROIs NET, ET, and edema; 1.1% were suspected of previous resection; 1.2% were suspected of a secondary disease; in 1.5% the FOV did not cover the whole brain; in 0.3% the registration among MR sequences was suboptimal.

## 1.3 Image Conversion & Tumor Segmentation

All selected DICOM series were converted to NIFTI images using the dicom2nifti[2] python library. Then, each patient's tumor was segmented using the automatic glioblastoma segmentation software DeepBraTumIA[3]. Provided T1, T1 contrast, T2 and FLAIR sequences as input, DeepBraTumIA identifies the enhancing tumor (ET), non-enhancing tumor (NET), edema subregions, and the overall brain tissue. Masks for each region-of-interest (ROI) are output in the co-registered reference of the MNI125 atlas[4], as well as in the frame of the respective original MR series. Exploiting the availability of brain masks and tumor regions of interest (ROIs), the original images were subjected to further pre-processing.

## 1.4 Intensity Standardization

As non-quantitative MR imaging lacks a physical/anatomic reference scale, direct quantitative comparison of image intensity across vendors and imaging protocols is not possible. Statistical standardization techniques provide a means to achieve comparability in the absence of such a reference by aligning the intensity distributions of anatomical reference structures across individual image acquisitions. In previous work [1], we compared the effect of different preprocessing strategies on MR-based radiomics prediction modeling. Our results indicated that bias-field correction applied to the entire images does not improve and may deteriorate prediction performance. Further, no significant difference could be shown between simple $z$-score standardization vs. white-stripe standardization, but standardization approaches that corrected for the presence of (patient-specific) tumor regions performed best. Building on this experience, no bias-field correction was applied in this study, and image intensities were aligned via $z$-score standardization and correcting for the presence of tumor regions: The mean $\mu$ and standard deviation $\sigma$ of intensity values $I_i$ was computed across all voxels $i$ of the healthy brain anatomy visible in a given image, i.e., excluding any non-brain regions and regions segmented as ET, NET, or edema. Image standardization then consisted in subtracting this mean intensity value from the intensity of each voxel and dividing the resulting intensity value by the standard deviation: $\hat{I}_i = (I_i - \mu)/\sigma$.

## 1.5 Image review

The processed images and segmentations were manually reviewed by an MR scientist with 10 years of experience and were further discussed with two radio-oncologists with 15 and 10 years of clinical experience, respectively. Only imaging exams that included all four MR contrasts and provided full brain coverage were considered for this study. Additionally, the data had to exhibit diagnostic image quality, characterized by visually acceptable image resolution, observable anatomical details, and the absence of artifacts or only minimal and normally occurring artifacts. Image registration across the four contrasts had to be accurate. We required tumor segmentations to include ROIs for the NET, ET, and edema components, and to be well performed according to visual assessment. MR data with extensive artifacts caused by motion, pulsation, folding, metal, or others, as well as data showing previous interventions such as tumor resection, or secondary diseases such as hydrocephalus or strokes, were excluded from further analysis.

---

[1] https://github.com/ccipd/MRQy
[2] https://github.com/icometrix/dicom2nifti
[3] https://www.nitrc.org/projects/deepbratumia/.
[4] https://www.mcgill.ca/bic/neuroinformatics/brain-atlases-human

| ROI name | Tumor subregion(s) |
|---|---|
| *ROI_1* | edema |
| *ROI_2* | enhancing tumor (ET) |
| *ROI_3* | non-enhancing tumor (NET) |
| *ROI_1-2* | edema & ET |
| *ROI_2-3* | ET & NET |
| *ROI_1-2-3* | ET & NET & edema |

Table 1: Tumor sub-regions (*ROI_1*, *ROI_2*, *ROI_3*) and combinations of subregions (*ROI_1-2*, *ROI_2-3*, *ROI_1-2-3*) from which radiomics features were extracted.

## 1.6 Radiomics Feature Extraction

In addition to the three tumor subregions provided by the automatic segmentation approach (NET, ET, edema), section 1.3, three further masks were created corresponding to the different combinations of those subregions that typically result in contiguous volumes (ET-edema, NET-ET, NET-ET-edema). Table 1 summarizes all ROIs considered in this study.

For CR-based modeling, radiomics feature sets comprising 107 shape, intensity, and texture features were computed from the standardized images using the pyradiomics [2] Python library. Extraction was performed from each MR contrast (T1, T1c, T2, FLAIR) and for each of the six (primary and combined) ROIs (NET, ET, edema, ET-edema, NET-ET, NET-ET-edema), resulting in a total of $6\,[\text{ROIs}] \times 4\,[\text{sequence types}] = 24$ feature sets per patient. The standardized images in their original frame of reference were resampled to a spatial resolution of $1\,\text{mm} \times 1\,\text{mm} \times 1\,\text{mm}$ prior to feature extraction. To ensure positive valued image intensities, we applied a shift of $+10$ to all normalized intensity values. The *bin_width* parameter for gray value discretization was set to a value of 0.3125, so that the full range of intensity values after standardization could be well captured by approximately 64 bins. For a small subset of patients, not all primary ROIs (NET, ET, edema) were identified by the automated segmentation progress, resulting in incomplete radiomics featuresets. To facilitate further analysis, these patients were excluded from modeling studies.

## 1.7 Co-registration for Deep Learning

While radiomics features were extracted from the standardized images in their original frame of references, for DL investigations, all standardized images were aligned by rigid registration to the MNI152 (T1, $1\,\text{mm} \times 1\,\text{mm} \times 1\,\text{mm}$) atlas[5]. For alignment, we applied the transformation matrices resulting from DeepBraTumIA's segmentation process to the standardized images and tumor segmentations. The transformed images/masks were resampled to the atlas resolution ($1\,\text{mm} \times 1\,\text{mm} \times 1\,\text{mm}$) using linear interpolation for images and nearest-neighbor interpolation for segmentation masks.

We determined the minimum useful field of view (FoV) by superposing all brain masks in the atlas frame. All co-registered images and segmentations were cropped to the minimum extent of the superposed brain masks plus a margin of $3\,\text{mm}$. Finally, the cropped co-registered standardized images were resampled to a matrix with isotropic dimensions of 160 voxels, corresponding to an isotropic spatial resolution of $1.2875\,\text{mm}$.

---

[5] https://nist.mni.mcgill.ca/icbm-152lin/

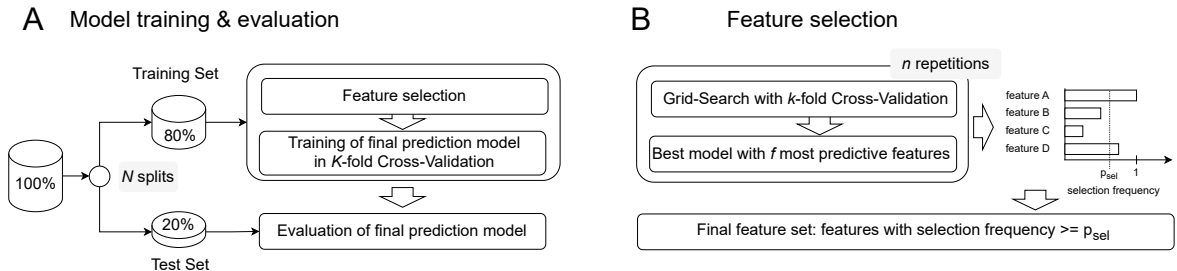

Figure 2: Conventional radiomics: Feature selection, model training, and (internal) evaluation.

## 2 Modeling Details

### 2.1 Conventional Radiomics Modeling

**Feature Selection** Most conventional radiomics modeling strategies seek to identify the subset of the most salient features from the large set of pre-computed quantitative image descriptors before model training. We used the three-step process, illustrated in fig. 2 (B), to achieve this: First, to identify a subset of predictive features, an optimal model with up to $f$ features was selected by $k$-fold grid-search cross-validation (CV) on the training data. This step was repeated $n$ times with different CV seeds to introduce variability in the datasets used for model training and evaluation in the CV, each time yielding an optimal model with up to $f$ (or fewer) features. In a second step, the selection frequency for each feature across all $n$ repetitions was computed. Features with a selection frequency $p_{sel}$ close to 1 were considered robust predictors for the entire training set, whereas features with $p_{sel}$ close to 0 were considered unlikely to provide robust predictive value. Finally, all features selected at least as frequently as the minimum selection frequency threshold $p_{sel}$ were included in the final feature set. Values for the hyperparameters $n$, $k$, and $p_{sel}$ were chosen based on prior experience: $n = 10$ and $k = 4$ had been found to provide a good compromise between capturing variability across different CV splits while ensuring sufficient data samples in the evaluation split. The minimum selection frequency threshold was set to $p_{sel} = 0.5$. The hyperparameter $f$ was optimized by repeating the selection process for several values of $f \in \{5, 10, 20, 30, 40, 50\}$ and selecting the value (and resulting features and models) that yielded the highest average CV training score across $N$ repetitions in the outer CV training loop, fig. 2 (A). In general, we found larger values of $f$ to increase the risk of overfitting, resulting in larger differences between CV and test performances.

**Models and Metrics** For survival problems, Cox Proportional Hazards models were used in training and evaluation with Harrel's Concordance Index (C-score) as performance metric; feature selection relied on Cox-Net models with combined $L1$ and $L2$ regularization. Classification problems relied on Logistic Regression models for training and evaluation, using the area under the receiver operator curve (ROC AUC) as performance metric; feature selection used Elastic-Net models.

**Batch Harmonization of Radiomics Feature Values** As the imaging data employed in this study was highly heterogeneous, we investigated the presence of systematic differences in radiomics feature values in function of device vendor, field strength, and data source by unsupervised clustering and exploration in 2D PCA-reduced feature space. Further, we explored the utility of statistical batch-harmonization techniques (ComBat) for mitigating the effect of covariate-specific differences on radiomics feature values, using the *neuroCombat* implementation[6]. The effect of ComBat batch normalization on model performance was not significant for neither 6-month survival nor overall survival endpoints. Compare table 3 in the manuscript (SMOTE, no batch normalization) with table 4 (right subtable: SMOTE with batch normalization) 6-month survival, and table 5 for OS prediction.

---

[6] http://github.com/Jfortin1/neuroCombat

**Synthetic Minority Class Oversampling in Training**   To compensate for the imbalance of 6-month survival outcomes in our dataset (22.9 to 25.2 % high-risk patients with OS $\leq$6 m, we employed the synthetic minority class oversampling (SMOTE) [3] technique during model training using the *imbalanced-learn* [7] implementation. The effect of SMOTE on model performance from 6-month survival prediction was found to be minor and not statistically significant, compare table 3 in the manuscript (SMOTE, no batch normalization) with table 4 (SMOTE vs no SMOTE, with and without batch normalization).

**Late Fusion**   Models including exclusively radiomics or clinical features were evaluated separately. We evaluated the joint predictive value of radiomics and clinical information via *late-fusion* by computing a single *radiomics risk score RRS* and including this score as an additional feature along with other clinical information in the model. The radiomics risk score summarizes the predictive image-derived information for each patient.

For conventional radiomics based on CoxPH or Logistic Regression models, the radiomics score $RRS_i$ for a specific patient $i$ is simply computed as the sum of the products of regression weight $\beta_j$ with patient-specific value $x_{i,j}$ of included feature $j$:

$$RRS_i = \sum_j \beta_j \, x_{i,j} \,. \tag{1}$$

**Model Training and Evaluation**   Feature selection, if applicable, and subsequent model training and validation was performed on 80% of the *TrainEval* cohort in a 4-fold CV; the resulting model was tested on 20% of the *TrainEval*. This process was repeated $N = 10$ times for statistical evaluation, resulting in $4 \times 10 = 40$ *CV* performance estimates and 10 *Internal Test* performance estimates. To validate prediction performance on an independent dataset, a new model was retrained on the entire *TrainEval* cohort using features selected more than 50% across these $N = 10$ repetitions. The performance of the resulting model was then evaluated on the unseen *ExtTest* cohort. The general training and evaluation scheme is detailed in Fig. 2 in the main manuscript.

## 2.2  Deep Learning based Modeling

Joint fusion neural networks for survival prediction were investigated, leveraging MRI data, tumor segmentation masks, and clinical parameters as inputs. We explored several well-established DL architectures, including ResNet and DenseNet with tailored modifications and hyperparameter optimizations (learning rate, depth, channels, kernel-size, dropouts, batch size, loss-function, and scheduler). Moreover, the 3D MRI data and segmentation were used as input on separate channels, or stacked on one single channel to leverage similar features.

The final neural network architecture is shown in Figure 3. It consists of three residual blocks for MR data processing and tumor segmentation, followed by two fully connected layers to include clinical information and demonstrated effectiveness while maintaining reduced computational cost. The network can be utilized for imaging-only, clinical-only, and combined imaging with clinical parameters. The first input to the network consists of four MR contrasts and the tumor multi-label segmentation mask, stacked into a single channel to form a unified input. This input leverages shared features between the raw MRI data and the segmentation and is processed through three residual blocks, each comprising two convolutional layers with skip connections. The convolutional blocks consist of a 3D convolution, group normalization and Leaky ReLU activation. Following the residual blocks, an adaptive max pooling is applied to flatten the features, and the clinical data (age, sex, MGMT, IDH) are jointly incorporated as the second input, just before the fully connected (FC) layer. The tensor is then processed by two FC layers consisting of linear transformation, group normalization, Leaky ReLU, and dropout. The final layer outputs probabilities for binary classification, indicating the likelihood of being in one category or another (e.g., surviving at 6 months or not). It is important to note that this approach does not involve survival analysis through risk score prediction or modeling the complete survival curve.

On-the-fly image data augmentation was used during training to diversify the input data and improve the robustness and generalization of the models. These include transformations such as rotation, shearing, translation, scaling, and grid distortion. Gaussian and Gibbs noise are employed to mimic noise commonly present in MRI, together with signal intensity modifications, including shifts and scaling. The data augmentation was fine-tuned to emulate potential variations in the imaging data while avoiding excessively pronounced and unrealistic transformations.

---

[7] http://imbalanced-learn.org

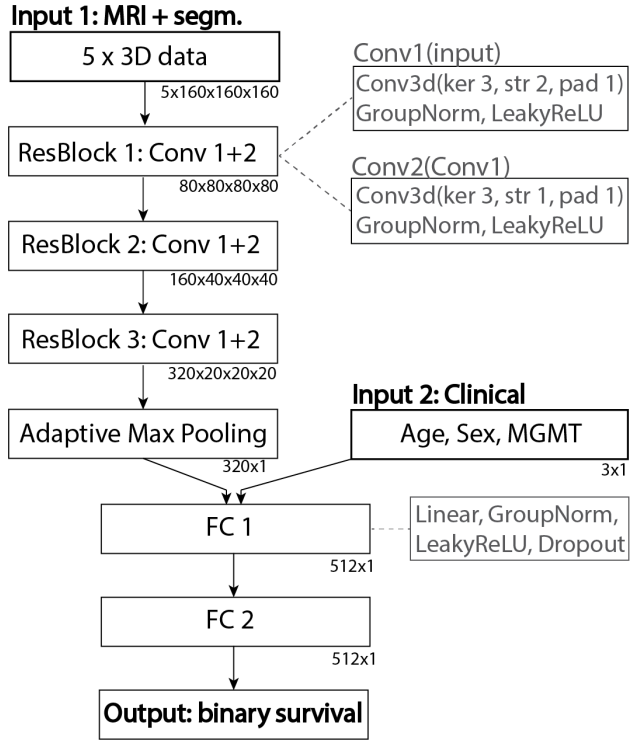

Figure 3: Architecture of the joint fusion DL model leveraging both MRI data and clinical parameters as inputs.

The cross-entropy loss function was employed, and the Adam optimizer initialized with a learning rate of $10^{-5}$ A cosine annealing warm restart learning rate scheduler was employed to cyclically adjust the learning rate, facilitating convergence. A batch size of 4 was used.

For each combination of imaging and clinical information, we trained ten models by shuffling the *TrainEval* cohort, and splitting it into 60-20-20 percentage for training, validation, and Internal Testing, as shown in **??**. We employed upsampling on the training data to address class imbalance. We trained every model for 24 hours and selected the model at the epoch with the highest AUC on the validation set for further analysis and evaluation on the Internal Testing, as well as on the *ExtTest* cohort. Additionally, for the *ExtTest* cohort, ensemble-averaged results (in terms of ROC-AUC) were computed from the averaged model prediction probabilities.

Pytorch[8] and Monai[9] libraries were used for the implementations. The code snippets of the DL architecture, data augmentation, and model parameters are provided in section 2.2.1.

### 2.2.1 Code

```
class ConvBlockOP(nn.Module):
    def __init__(self, in_channels, out_channels, kernel_size, stride, padding,
        ↪ dropout_rate, num_groups):
        super(ConvBlockOP, self).__init__()
```

---
[8]http://pytorch.org/
[9]http://monai.io/

```python
        self.conv = nn.Conv3d(in_channels, out_channels, kernel_size, stride, padding)
        self.gn = nn.GroupNorm(num_groups=num_groups, num_channels=out_channels)
        self.relu = nn.LeakyReLU(negative_slope=0.1)
        self.dropout = nn.Dropout3d(dropout_rate)

    def forward(self, x):
        x = self.conv(x)
        x = self.gn(x)
        x = self.relu(x)
        x = self.dropout(x)
        return x

class ResBlockOP(nn.Module):
    def __init__(self, in_channels, out_channels):
        super(ResBlockOP, self).__init__()
        self.conv1 = ConvBlockOP(in_channels, out_channels, kernel_size=3, stride=2,
            ↪ padding=1, dropout_rate=0.00, num_groups=8)
        self.conv2 = ConvBlockOP(out_channels, out_channels, kernel_size=3, stride=1,
            ↪ padding=1, dropout_rate=0.00, num_groups=8)

    def forward(self, x):
        x = self.conv1(x)
        x2 = self.conv2(x)
        x = x + x2
        return x

class ClassifierOP(nn.Module):
    def __init__(self, in_channels, num_classes):
        super(ClassifierOP, self).__init__()
        self.resblock1 = ResBlockOP(in_channels, 80)
        self.resblock2 = ResBlockOP(80, 160)
        self.resblock3 = ResBlockOP(160, 320)
        self.max_pool =nn.AdaptiveMaxPool3d((1,1,1))
        self.dropout0=nn.Dropout(p=0.0)
        self.fc1 = nn.Linear(320+3, 512)
        self.bn1 = nn.GroupNorm(num_groups=8,num_channels=512)
        self.dropout1=nn.Dropout(p=0.5)
        self.fc2 = nn.Linear(512, 512)
        self.bn2 = nn.GroupNorm(num_groups=8, num_channels=512)
        self.dropout2=nn.Dropout(p=0.5)
        self.fc3 = nn.Linear(512, 2)
        self.relu = nn.LeakyReLU(negative_slope=0.1)

    def forward(self, x, mgmt, age, sex, idh):
        mgmt = torch.tensor(mgmt, dtype=torch.float32).unsqueeze(1)
        age = torch.tensor(age, dtype=torch.float32).unsqueeze(1)
        sex = torch.tensor(sex, dtype=torch.float32).unsqueeze(1)

        x = self.resblock1(x)
        x = self.resblock2(x)
        x = self.resblock3(x)

        x = self.max_pool(x)
        x = x.view(x.size(0), -1)
```

```python
        x = self.dropout0(x)

        x = self.fc1(torch.cat((x, mgmt, age, sex,idh), dim=1))
        x = self.bn1(x)
        x = self.relu(x)
        x = self.dropout1(x)

        x = self.fc2(x)
        x = self.bn2(x)
        x = self.relu(x)
        x = self.dropout2(x)

        x = self.fc3(x)
        return x

train_transforms_image = Compose([EnsureChannelFirst(),
    RandAffine(prob=0.25,padding_mode='zeros',rotate_range=(np.pi / 25, np.pi / 25, np.pi
        ↪ / 25), mode="bilinear"),
    RandAffine(prob=0.25,padding_mode='zeros',shear_range=(0.11,0.11,0.11), mode="
        ↪ bilinear"),
    RandAffine(prob=0.25,padding_mode='zeros',translate_range=(12,12,12), mode="bilinear"
        ↪ ),
    RandAffine(prob=0.25,padding_mode='zeros',scale_range=(0.2, 0.2, 0.2), mode="bilinear
        ↪ "),
    RandGridDistortion(num_cells=3, prob=0.1, distort_limit=(-0.025, 0.025),padding_mode=
        ↪ 'zeros',mode="bilinear"),
    # RandFlip(prob=0.5, spatial_axis=0),
    RandGaussianNoise(prob=0.1, mean=0.0, std=0.015),
    RandShiftIntensity(0.05, prob=0.1),
    RandScaleIntensity(0.05, prob=0.1),
    RandGibbsNoise(prob=0.1, alpha=(0.0,0.2)),
    EnsureType()])

train_transforms_segmentation = Compose([EnsureChannelFirst(),
    RandAffine(prob=0.25,padding_mode='zeros',rotate_range=(np.pi / 25, np.pi / 25, np.pi
        ↪ / 25), mode="nearest"),
    RandAffine(prob=0.25,padding_mode='zeros',shear_range=(0.11,0.11,0.11), mode="nearest
        ↪ "),
    RandAffine(prob=0.25,padding_mode='zeros',translate_range=(12,12,12), mode="nearest")
        ↪ ,
    RandAffine(prob=0.25,padding_mode='zeros',scale_range=(0.2, 0.2, 0.2), mode="nearest"
        ↪ ),
    RandGridDistortion(num_cells=3, prob=0.1, distort_limit=(-0.025, 0.025),padding_mode=
        ↪ 'zeros',mode="nearest"),
    # RandFlip(prob=0.5, spatial_axis=0),
    EnsureType()])

loss_function = torch.nn.CrossEntropyLoss()
optimizer = torch.optim.Adam(model.parameters(), lr)
batch_size=4
model = ClassifierOP(in_channels=5, num_classes=2). to(device)
lr=1e-5
```

```
scheduler = CosineAnnealingWarmRestarts(optimizer,
    T_0 = 10, # Number of iterations for the first restart
    T_mult = 1, # A factor increases T_0 after a restart (none)
    eta_min = 1e-6) # Minimum learning rate
```

Listing 1: Code snippets of the DL architecture, data augmentation, and model parameters: Pytorch and Monai libraries were used for implementation.

# 3 Additional Results

## 3.1 Clinical Characteristics and Survival Characteristics

As complement to Table 2 of the manuscript, table 2 summarizes the clinical characteristics of all patients included in this modeling study, organized by data source. Table 3 summarizes the results of statistical testing for differences in clinical covariates in *TrainEval* vs. *ExtTest* cohorts, across all data subsets *S-1* to *S-3*. No statistically significant differences were found between the distributions of clinical covariates (age, sex, mgmt status) and outcome (6-months survival) in the *TrainEval* and *ExtTest* cohort for all data subsets except for patient age in subset *S-1* ($p = 0.001$).

| Data Origin | Patients | Age | | | Sex | MGMT | | Survival | | |
|---|---|---|---|---|---|---|---|---|---|---|
| | | mean | median | range | female | meth | NA | mean | median | < 6m |
| | # | [y] | [y] | [y] | [%] | [%] | [%] | [d] | [d] | [%] |
| All Centers | 1152 | 64.1 | 65.0 | [19,90] | 36.2 | 41.2 | 32.3 | 465.3 | 380.0 | 24.5 |
| UPENN | 418 | 63.7 | 64.0 | [21,64] | 37.6 | 34.8 | 51.9 | 427.9 | 370.0 | 27.3 |
| SPHN-1 | 383 | 63.8 | 65.0 | [19,65] | 36.0 | 40.2 | 37.1 | 490.6 | 380.0 | 22.5 |
| SPHN-2 | 81 | 60.9 | 61.0 | [25,61] | 34.6 | 50.6 | 0.0 | 558.3 | 437.0 | 17.3 |
| SPHN-3 | 245 | 66.5 | 67.0 | [25,67] | 34.7 | 45.0 | 2.9 | 466.3 | 398.0 | 24.1 |
| SPHN-4 | 16 | 65.3 | 62.0 | [32,63] | 37.5 | 38.5 | 18.8 | 348.3 | 295.5 | 43.8 |
| SPHN-5 | 9 | 63.0 | 65.0 | [37,65] | 33.3 | 16.7 | 33.3 | 462.2 | 374.0 | 22.2 |

Table 2: Patient characteristics by data origin. See table 2 of the manuscript for a comparison of patient characteristics by cohorts (*TrainEval* vs. *ExtTest*) and data subsets. Values in the MGMT columns refer to the percentage of patients with MGMT hyper-methylation (*meth*) and unknown (*NA*) MGMT status, respectively.

| Subset | Age | Sex | MGMT | 6m survival | survival |
|---|---|---|---|---|---|
| | Mann-Whitney-U | prop. $z$-test | prop. $z$-test | prop. $z$-test | log-rank test |
| *S-1* | 0.001 | 0.59 | – | 0.76 | 0.72 |
| *S-2* | 0.14 | 0.96 | 0.26 | 0.71 | 0.88 |
| *S-3* | 0.35 | 0.86 | 0.13 | 0.74 | 0.61 |

Table 3: Statistical tests and corresponding *p*-values for comparing the distribution of clinical variables across *TrainEval* and *ExtTest* cohorts for all three data subsets *S-1*-*S-3*. Only the distribution of patients' age in subset *S-1* was found to differ significantly ($p<0.05$) across the cohorts.

Figure 4 illustrates the overall-survival patterns: Figure 4a by cohort (*TrainEval*, *ExtTest*) across data subsets, and figs. 4b to 4d by data source and for each data subset, *S-1* to *S-3*, respectively.

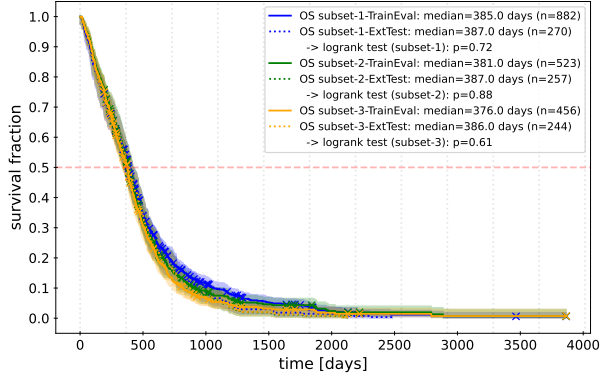

(a) By cohort (*TrainEval*, *ExtTest*) across data subsets (*S-1*, *S-2*, *S-3*)

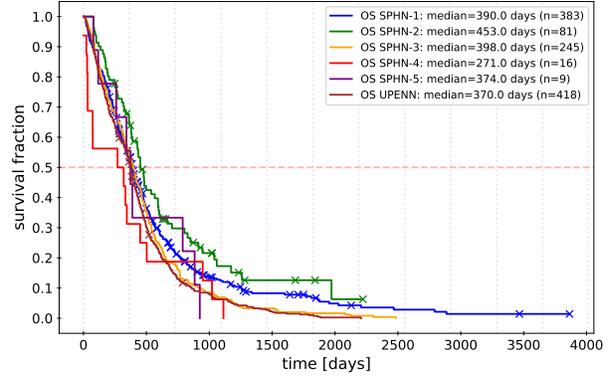

(b) By center for data subset *S-1*.

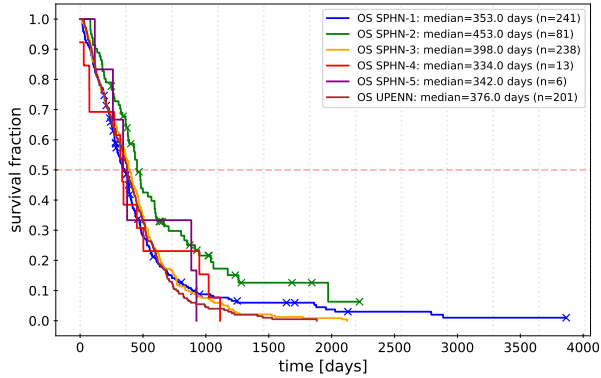

(c) By center for data subset *S-2*.

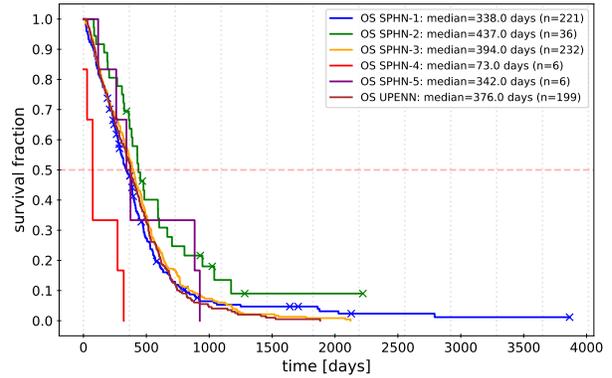

(d) By center for data subset *S-3*.

Figure 4: Kaplan-Meier survival curves for all data subsets and cohorts. Subfig. (a) compares the survival characteristics of *TrainEval* and *ExtTest* cohorts for all data subsets. No statistically significant differences were found between the survival characteristics of both cohorts within any given data subset (log-rank test $p > 0.05$). Subfigs. (b)-(d) show the survival characteristics by data origin for data subsets *S-1*, *S-2* and *S-3*, respectively.

## 3.2 Additional CR Results for 6-month Survival Prediction

Figure 5 compares the distribution of AUC values (6-month survival) obtained for CR (SMOTE, no batch harmonization) and DL modeling approaches using the different data subsets, corresponding to the results summarized in table 3 of the manuscript.

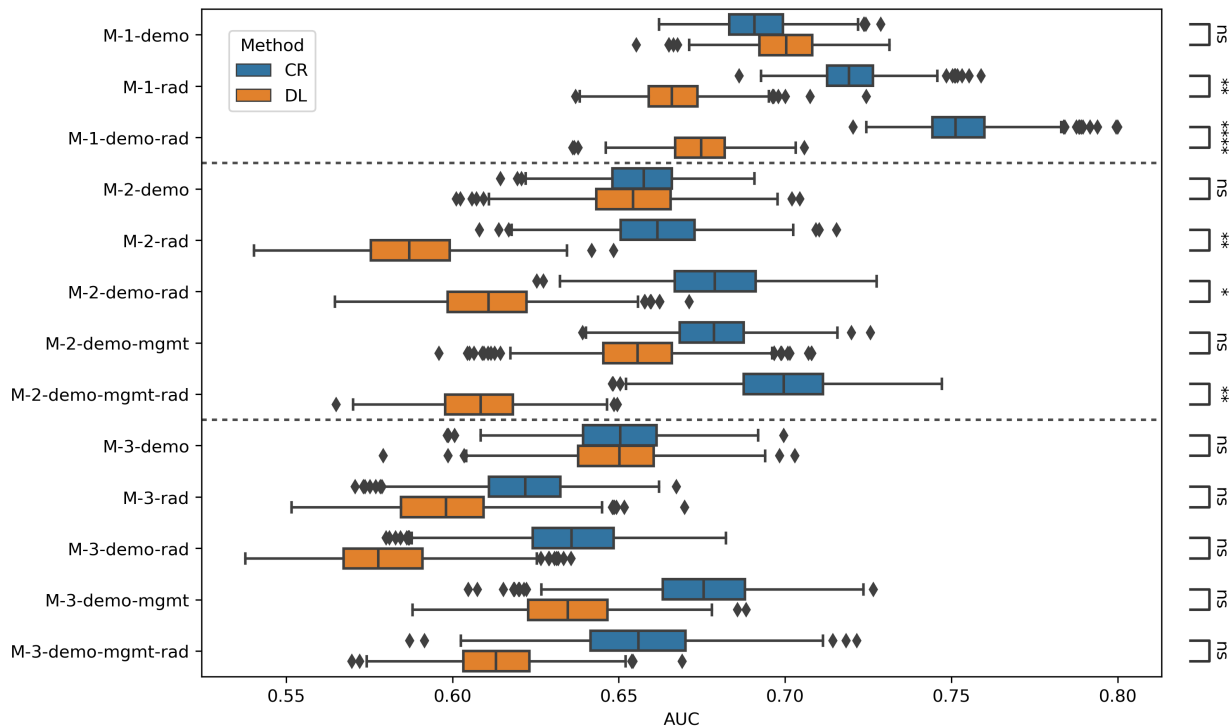

Figure 5: Average AUC performance for 6-month survival prediction on *TrainEval* (statistical evaluation on Internal Test set, table 3 in manuscript) from bootstrapping ($n = 1000$). Average performance of CR and DL approaches was compared via Permutation Test with significance levels: *ns* ($p > 0.05$), * ($p \leq 0.05$), ** ($p \leq 0.01$), *** ($p \leq 0.001$), and **** ($p \leq 0.0001$).

Figure 6 compares the permutation test *p*-values resulting from pair-wise comparison among all CR and DL models for 6-months survival prediction. In the CR approach, model *M1-demo-img* performed significantly better ($p = 0.002$) than model *M1-demo*, relying exclusively on demographic information. This could not be replicated by the DL modeling approach.

Section 3.2 compares the Odds-Ratio (OR) of selected features for 6-month survival in *TrainEval* and *ExtTest* cohorts for all data subsets *S-1* to *S-3* and CR models. Both *age* and *RRS* (radiomics_score) are significantly associated with 6-month survival across data subsets, cohorts and independent of other variables.

Figure 8 illustrates the ROC curves and optimal decision thresholds for DL and CR 6-months survival models on subcohort *S-1*.

Figure 9 summarizes the features selected for the final CR 6-months survival prediction model evaluated on *extTest*. See table 1 for the naming ROIs.

Figure 10 illustrates the models' ability to distinguish short vs. medium-long survivors. All models (clinical-only, radiomics-only, combined) stratify patients into cohorts with significantly different survival.

Table 4 summarizes ROC-AUC performances for alternative CR modeling strategies: (a) without use of synthetic minority oversampling technique (SMOTE) and without feature-based batch harmonization (ComBat), (b) employing both SMOTE and ComBat. To conceptually align with the DL modeling strategy, the CR model presented in table 3 of the manuscript employed SMOTE, but did not make use of batch harmonization. No statistically significant differences were observed between the test performance on the

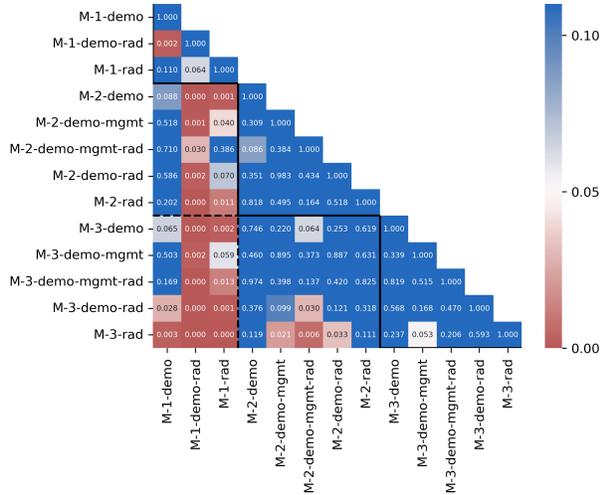 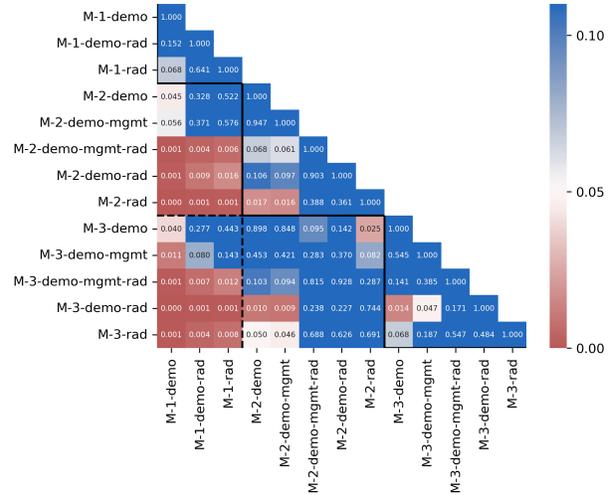

(a) Model comparison for CR.  (b) Model comparison for DL.

Figure 6: Permutation test $p$-values indicating differences in the *mean AUC* (6-month survival prediction) between all pairs of models and subcohorts for CR (a) and DL (b), respectively. Pairs of models with p-values $p < 0.05$ are considered to differ *significantly*.

*TrainEval* cohorts across these modeling strategies. Point estimates on the *ExtTest* dataset were found to differ only slightly by $\leq \pm 0.02$ AUC.

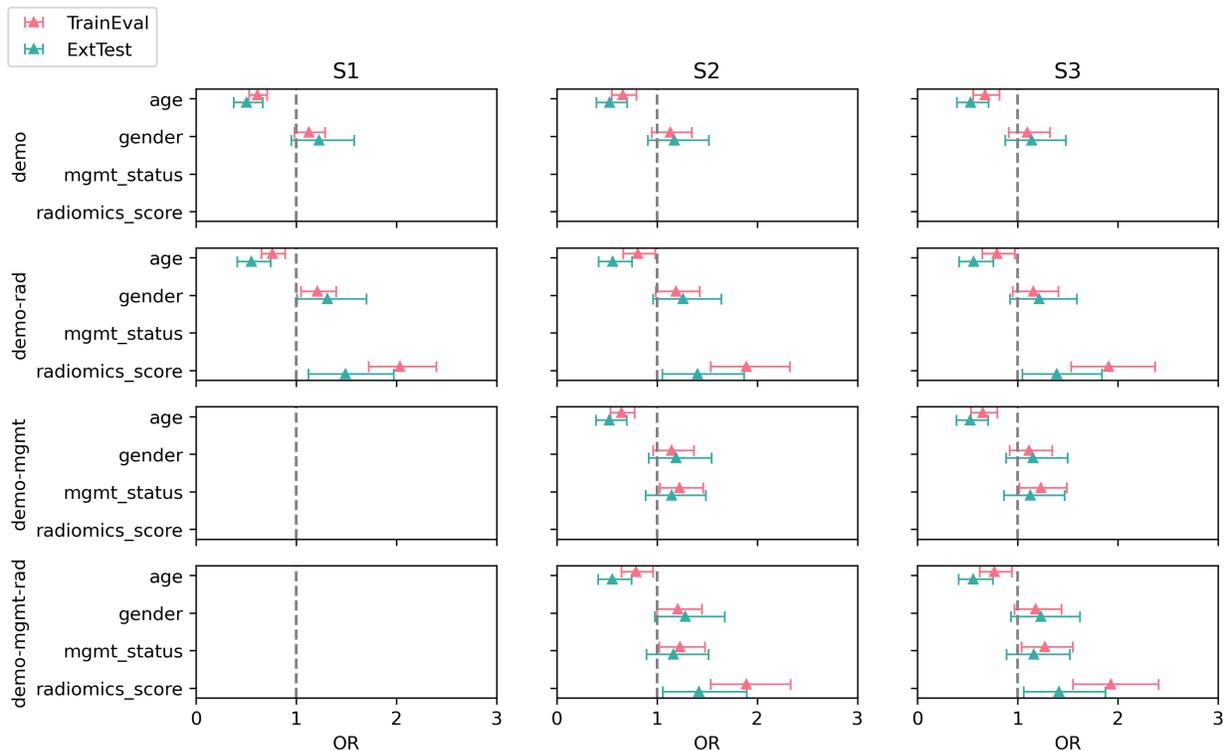

Figure 7: Comparison of Odds-Ratio (OR) of selected features (*age*, *gender*, *MGMT*-status and radiomics score *RRS*) for 6-month survival in *TrainEval* and *ExtTest* cohorts for all data subsets *S-1* to *S-3* and models.

|  |  | **CR (no SMOTE, no batch norm.)** | | | **CR (SMOTE, batch norm.)** | | |
|---|---|---|---|---|---|---|---|
|  |  | CV-Validation | Internal Test | Test | CV-Validation | Internal Test | Test |
| *S-1* | *M1-demo* | 0.70 [0.68,0.71] | 0.70 [0.67,0.72] | 0.74 | 0.69 [0.68,0.70] | 0.69 [0.67,0.72] | 0.74 |
|  | *M1-img* | 0.76 [0.75,0.77] | 0.73 [0.71,0.75] | 0.68 | 0.76 [0.75,0.77] | 0.72 [0.70,0.74] | 0.69 |
|  | *M1-demo-img* | 0.79 [0.78,0.80] | 0.76 [0.74,0.78] | 0.75 | 0.80 [0.78,0.81] | 0.75 [0.73,0.77] | 0.76 |
| *S-2* | *M2-demo* | 0.68 [0.67,0.70] | 0.66 [0.63,0.68] | 0.73 | 0.68 [0.66,0.70] | 0.66 [0.63,0.68] | 0.73 |
|  | *M2-img* | 0.78 [0.76,0.80] | 0.65 [0.62,0.67] | 0.66 | 0.76 [0.75,0.78] | 0.65 [0.61,0.69] | 0.66 |
|  | *M2-demo-img* | 0.83 [0.81,0.84] | 0.67 [0.64,0.69] | 0.71 | 0.80 [0.78,0.81] | 0.68 [0.64,0.72] | 0.71 |
|  | *M2-demo-mgmt* | 0.69 [0.67,0.71] | 0.68 [0.65,0.71] | 0.73 | 0.69 [0.67,0.70] | 0.68 [0.65,0.71] | 0.72 |
|  | *M2-demo-mgmt-img* | 0.84 [0.82,0.85] | 0.69 [0.66,0.71] | 0.73 | 0.80 [0.79,0.82] | 0.69 [0.65,0.74] | 0.73 |
| *S-3* | *M3-demo* | 0.67 [0.66,0.69] | 0.65 [0.62,0.68] | 0.73 | 0.67 [0.65,0.69] | 0.65 [0.62,0.68] | 0.73 |
|  | *M3-img* | 0.78 [0.76,0.79] | 0.62 [0.59,0.64] | 0.64 | 0.76 [0.75,0.78] | 0.61 [0.57,0.65] | 0.64 |
|  | *M3-demo-img* | 0.83 [0.81,0.85] | 0.64 [0.60,0.67] | 0.69 | 0.80 [0.78,0.82] | 0.63 [0.59,0.68] | 0.70 |
|  | *M3-demo-mgmt* | 0.68 [0.66,0.70] | 0.67 [0.64,0.71] | 0.72 | 0.67 [0.65,0.69] | 0.67 [0.63,0.71] | 0.71 |
|  | *M3-demo-mgmt-img* | 0.84 [0.82,0.85] | 0.66 [0.62,0.69] | 0.72 | 0.81 [0.79,0.83] | 0.66 [0.61,0.71] | 0.71 |

Table 4: Comparison of ROC-AUC performances for 6m-survival conventional radiomics models without (left) and with (right) synthetic minority outcome class oversampling (SMOTE) and batch harmonization (ComBat) of radiomics features. No statistically significant differences were observed between the test performance on the *TrainEval* cohort between these two approaches, and the one presented in table 3 of the manuscript (SMOTE, no batch normalization).

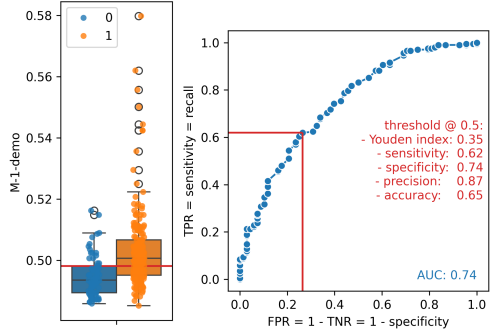
(a) ROC-demo-DL-S1

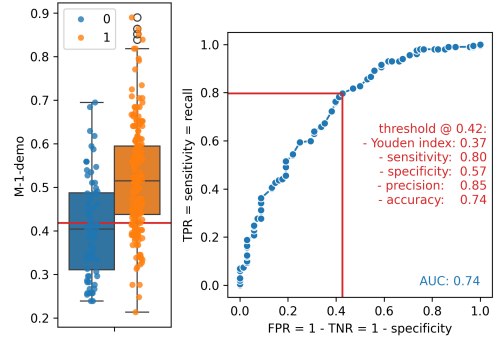
(b) ROC-demo-CR-S1

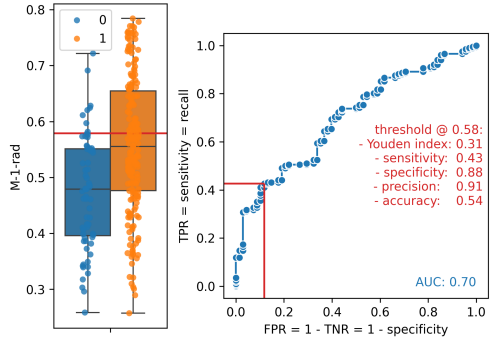
(c) ROC-img-DL-S1

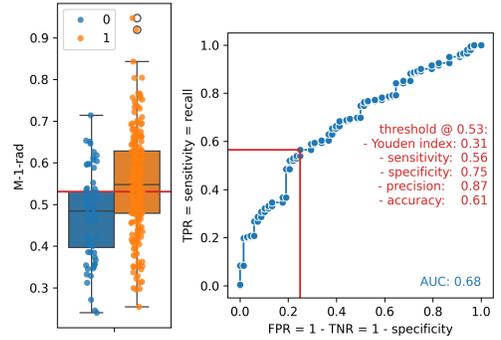
(d) ROC-demo-CR-S1

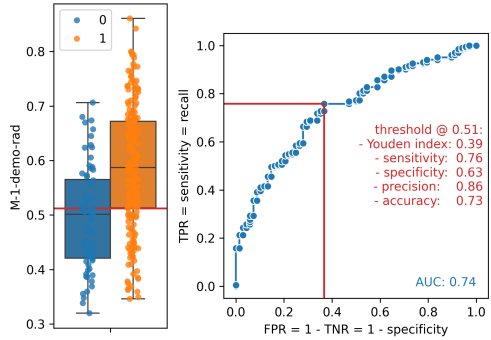
(e) ROC-img-clinical-DL-S1

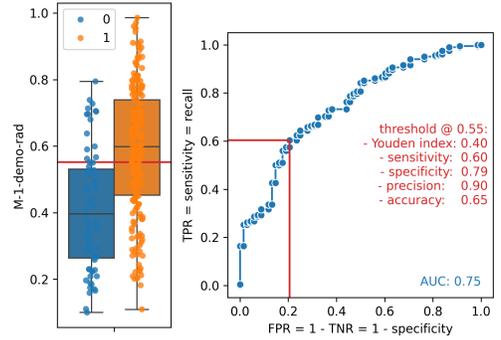
(f) ROC-demo-CR-S1

Figure 8: Receiver-Operator Curve and optimal decision thresholds for the DL (left) and CR (right) model on the entire cohort (Subset *S-1*). Each plot shows the distribution of image-derived scores for patients with OS <6 m (blue) and OS ≥6 m (orange) on the left and the ROC curve on the right. Red lines indicate the optimal decision threshold for classification according to Youden.

| feature | M-1-rad | M-2-rad | M-3-rad |
|---|---|---|---|
| All/ROI_1-2-3/shape_Maximum2DDiameterSlice | TRUE | FALSE | FALSE |
| All/ROI_1-2/shape_Sphericity | TRUE | TRUE | TRUE |
| All/ROI_2/shape_Flatness | TRUE | TRUE | TRUE |
| All/ROI_2/shape_Sphericity | TRUE | TRUE | TRUE |
| All/ROI_3/shape_Sphericity | TRUE | FALSE | TRUE |
| FLAIR/ROI_1-2-3/glcm_Imc1 | FALSE | FALSE | TRUE |
| T1/ROI_1/firstorder_10Percentile | TRUE | FALSE | FALSE |
| T1/ROI_1/ngtdm_Contrast | TRUE | FALSE | FALSE |
| T1/ROI_1/ngtdm_Strength | TRUE | TRUE | TRUE |
| T1/ROI_2/gldm_LargeDependenceLowGrayLevelEmphasis | TRUE | TRUE | FALSE |
| T1/ROI_3/firstorder_90Percentile | TRUE | TRUE | TRUE |
| T1/ROI_3/firstorder_Skewness | FALSE | TRUE | TRUE |
| T1/ROI_3/glszm_LowGrayLevelZoneEmphasis | FALSE | FALSE | TRUE |
| T1c/ROI_1-2/firstorder_Median | FALSE | TRUE | FALSE |
| T1c/ROI_1-2/glcm_ClusterShade | FALSE | TRUE | TRUE |
| T1c/ROI_1-2/gldm_DependenceVariance | TRUE | FALSE | FALSE |
| T1c/ROI_1-2/gldm_LargeDependenceEmphasis | FALSE | TRUE | TRUE |
| T1c/ROI_1-2/glrlm_LongRunEmphasis | FALSE | FALSE | TRUE |
| T1c/ROI_1/firstorder_10Percentile | FALSE | TRUE | TRUE |
| T1c/ROI_1/glcm_Idmn | TRUE | FALSE | FALSE |
| T1c/ROI_2/glcm_Imc2 | TRUE | FALSE | FALSE |
| T2/ROI_1-2-3/firstorder_90Percentile | FALSE | TRUE | FALSE |
| T2/ROI_1-2-3/glrlm_RunLengthNonUniformityNormalized | FALSE | FALSE | TRUE |
| T2/ROI_1-2-3/glrlm_ShortRunEmphasis | FALSE | TRUE | TRUE |
| T2/ROI_1-2-3/ngtdm_Busyness | TRUE | TRUE | TRUE |
| T2/ROI_1-2/glcm_Idmn | FALSE | FALSE | TRUE |
| T2/ROI_1/firstorder_10Percentile | TRUE | FALSE | FALSE |
| T2/ROI_1/glcm_Idmn | TRUE | TRUE | TRUE |
| T2/ROI_1/gldm_DependenceNonUniformityNormalized | FALSE | TRUE | FALSE |
| T2/ROI_1/ngtdm_Contrast | TRUE | FALSE | FALSE |
| T2/ROI_3/firstorder_10Percentile | TRUE | TRUE | TRUE |
| T2/ROI_3/firstorder_Mean | FALSE | TRUE | TRUE |
| T2/ROI_3/firstorder_RootMeanSquared | FALSE | TRUE | TRUE |
| T2/ROI_3/firstorder_Skewness | TRUE | FALSE | FALSE |
| T2/ROI_3/glcm_Autocorrelation | TRUE | FALSE | FALSE |
| T2/ROI_3/gldm_HighGrayLevelEmphasis | TRUE | FALSE | FALSE |
| T2/ROI_3/ngtdm_Busyness | TRUE | FALSE | FALSE |

Figure 9: Radiomics features selected for final CR 6-months survival prediction model evaluated on *extTest* set, for subsets *S-1*, *S-2* and *S-3*, respectively.

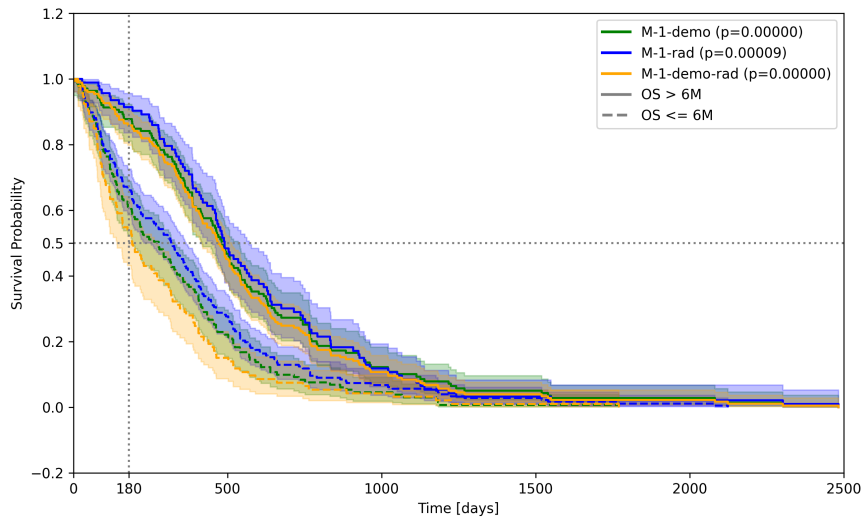

Figure 10: Kaplan-Meier survival curves of short vs. medium-long survivors based on CR model predictions in subcohort *S-1*. All models (clinical-only, radiomics-only, combined) stratify patients into cohorts with significantly different survival.

## 3.3 Additional CR Results for Overall Survival Prediction

Table 5 summarizes C-index performances for Overall-Survival (OS) modeling strategies with and without feature batch normalization.

Figure 11 compares the permutation test *p*-values resulting from pair-wise comparison among all CR models for OS prediction.

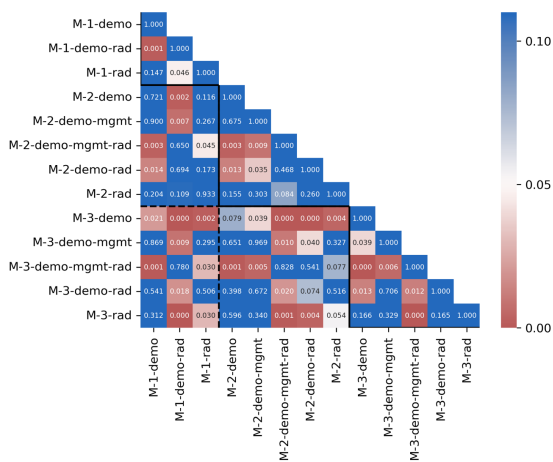

Figure 11: Permutation test *p*-values indicating differences in the *mean C-index* (OS prediction) between all pairs of models and subcohorts for CR models. Pairs of models with p-values $p < 0.05$ are considered to differ *significantly*

Figure 12 compares the Hazards-Ratio (HR) of selected features for OS survival in *TrainEval* and *ExtTest* cohorts for all data subsets *S-1* to *S-3* and CR models: *age*, *RRS* (radiomics_score) and *MGMT* are significantly associated with OS across data subsets and cohorts.

|  | | CR (no SMOTE, no batch norm.) | | | CR (no SMOTE, batch norm.) | | |
|--|--|---|---|---|---|---|---|
|  | | CV-Validation | Internal Test | Test | CV-Validation | Internal Test | Test |
| S-1 | M1-demo | 0.63 [0.62,0.64] | 0.64 [0.63,0.66] | 0.64 | 0.63 [0.63,0.64] | 0.64 [0.62,0.65] | 0.64 |
| S-1 | M1-img | 0.67 [0.67,0.68] | 0.66 [0.64,0.67] | 0.59 | 0.65 [0.63,0.66] | 0.65 [0.64,0.67] | 0.61 |
| S-1 | M1-demo-img | 0.70 [0.69,0.70] | 0.68 [0.67,0.69] | 0.63 | 0.69 [0.69,0.70] | 0.68 [0.66,0.69] | 0.64 |
| S-2 | M2-demo | 0.62 [0.62,0.63] | 0.64 [0.62,0.66] | 0.63 | 0.62 [0.61,0.63] | 0.64 [0.62,0.66] | 0.63 |
| S-2 | M2-img | 0.66 [0.65,0.67] | 0.66 [0.64,0.68] | 0.61 | 0.65 [0.64,0.66] | 0.65 [0.63,0.67] | 0.63 |
| S-2 | M2-demo-img | 0.68 [0.68,0.69] | 0.68 [0.66,0.69] | 0.64 | 0.68 [0.67,0.69] | 0.67 [0.65,0.68] | 0.65 |
| S-2 | M2-demo-mgmt | 0.65 [0.64,0.66] | 0.64 [0.63,0.66] | 0.64 | 0.65 [0.64,0.66] | 0.65 [0.64,0.67] | 0.64 |
| S-2 | M2-demo-mgmt-img | 0.70 [0.69,0.71] | 0.69 [0.67,0.71] | 0.66 | 0.70 [0.69,0.71] | 0.69 [0.67,0.71] | 0.67 |
| S-3 | M3-demo | 0.61 [0.60,0.62] | 0.62 [0.60,0.63] | 0.63 | 0.61 [0.60,0.62] | 0.62 [0.60,0.64] | 0.63 |
| S-3 | M3-img | 0.65 [0.64,0.66] | 0.63 [0.62,0.65] | 0.62 | 0.66 [0.66,0.67] | 0.63 [0.61,0.64] | 0.59 |
| S-3 | M3-demo-img | 0.68 [0.67,0.69] | 0.65 [0.63,0.67] | 0.65 | 0.67 [0.66,0.69] | 0.64 [0.62,0.66] | 0.63 |
| S-3 | M3-demo-mgmt | 0.63 [0.63,0.64] | 0.65 [0.63,0.66] | 0.63 | 0.63 [0.62,0.64] | 0.64 [0.63,0.66] | 0.63 |
| S-3 | M3-demo-mgmt-img | 0.69 [0.69,0.70] | 0.68 [0.67,0.70] | 0.67 | 0.69 [0.68,0.70] | 0.67 [0.65,0.69] | 0.66 |

Table 5: Comparison of C-Index performances for overall-survival CR models (no SMOTE, with and without batch normalization).

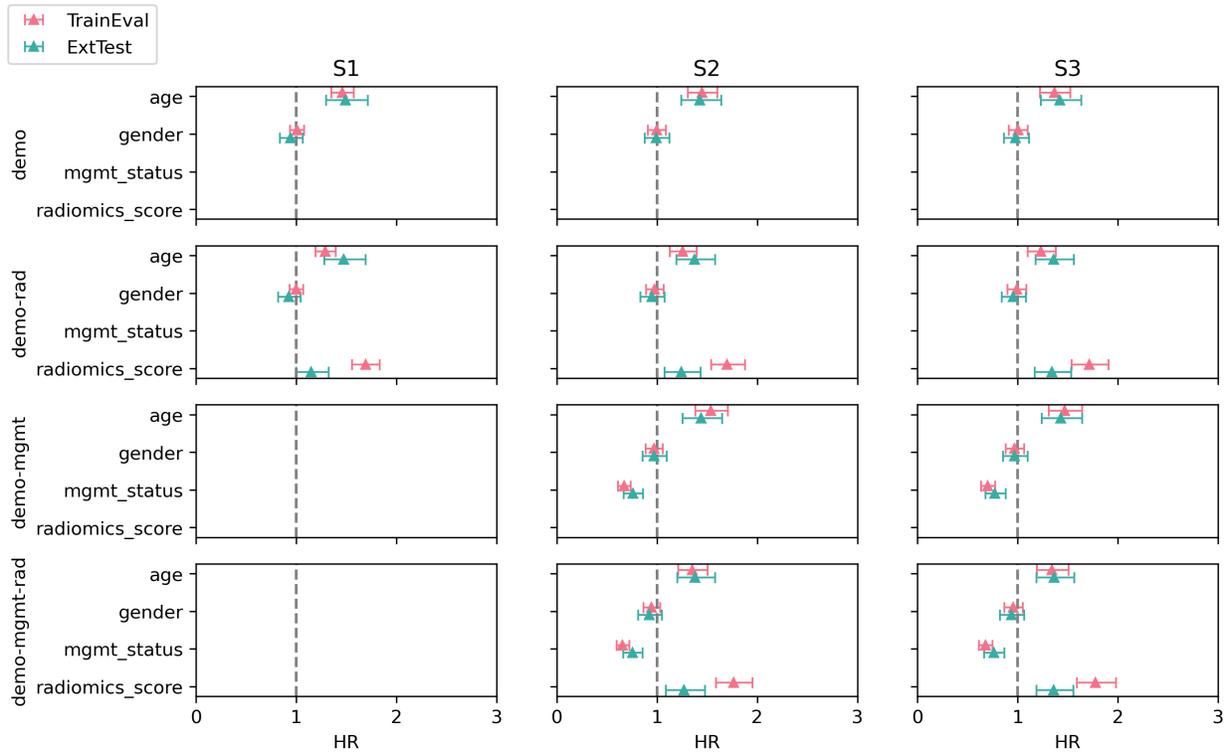

Figure 12: Comparison of Hazard-Ratio (HR) of selected features (*age*, *gender*, *MGMT*-status and radiomics score *RRS*) for overall survival in *TrainEval* and *ExtTest* cohorts for all data subsets *S-1* to *S-3* and models.